\documentclass[10pt,twocolumn,letterpaper]{article}
\pdfoutput=1
\usepackage{iccv}
\usepackage{times}
\usepackage{epsfig}
\usepackage{graphicx}
\usepackage{amsmath}
\usepackage{amssymb}

\usepackage{multirow}

\usepackage{caption}
\usepackage{booktabs}

\newcommand\blfootnote[1]{%
  \begingroup
  \renewcommand\thefootnote{}\footnote{#1}%
  \addtocounter{footnote}{-1}%
  \endgroup
}

\usepackage[pagebackref=true,breaklinks=true,colorlinks,bookmarks=false]{hyperref}

\iccvfinalcopy 

\def\iccvPaperID{****} 

\ificcvfinal\fi

\begin{document}

\title{ \vspace{-0.6cm} Make Encoder Great Again in 3D GAN Inversion through Geometry and Occlusion-Aware Encoding} 
\vskip -0.1 in
\author{%
	Ziyang Yuan$^{1*}$ \quad
    Yiming Zhu$^{1,2*}$\quad
    Yu Li$^{2\dagger}$ \quad
    Hongyu Liu$^{3}$ \quad
    Chun Yuan$^{1\dagger}$ \quad \\ 
    \vspace{-0.05cm}
    $^{1}$Tsinghua Shenzhen International Graduate School \quad $^{2}$International Digital Economy Academy (IDEA)\\
    $^{3}$Hong Kong University of Science and Technology \\
    \small \url{https://eg3d-goae.github.io}
}
\vspace{-0.6cm}

\twocolumn[{%
    \renewcommand\twocolumn[1][]{#1}%
    \vspace{-0.5cm}
    \maketitle
    \begin{center}
        \centering
        \vspace{-0.5cm}
        \includegraphics[width=0.95\textwidth]{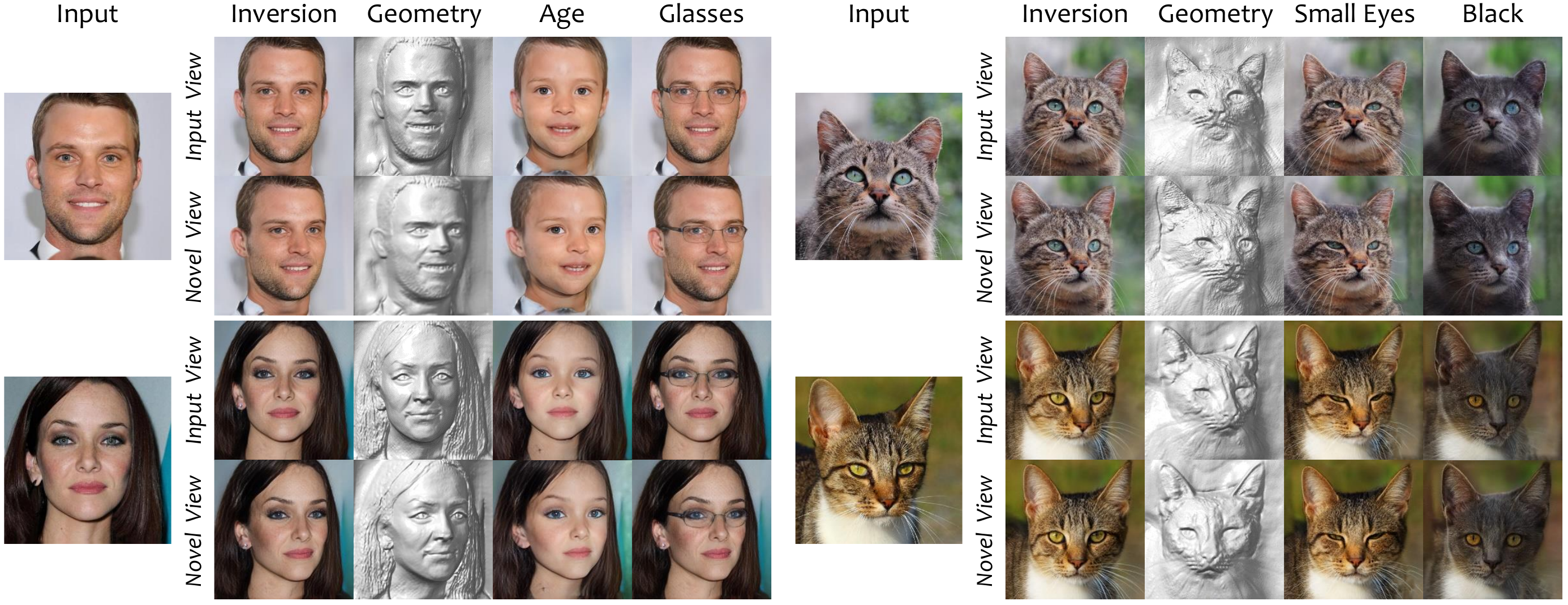}
        \captionof{figure}{\textbf{ The inversion and editing results of our method on human portraits and cat images.} The novel view synthesized images demonstrate that our method can achieve high-quality inversion and editing given a single image input.}
        \label{fig:teaser}
        \vspace{-0cm}
    \end{center}
}]

\ificcvfinal\thispagestyle{empty}\fi

\begin{abstract}
\vspace{-0.4cm}
    3D GAN inversion aims to achieve high reconstruction fidelity and reasonable 3D geometry simultaneously from a single image input. However, existing 3D GAN inversion methods rely on time-consuming optimization for each individual case. In this work, we introduce a novel encoder-based inversion framework based on EG3D, one of the most widely-used 3D GAN models. We leverage the inherent properties of EG3D's latent space to design a  discriminator and a background depth regularization. This enables us to train a geometry-aware encoder capable of converting the input image into corresponding latent code. Additionally, we explore the feature space of EG3D and develop an adaptive refinement stage that improves the representation ability of features in EG3D to enhance the recovery of fine-grained textural details. Finally, we propose an occlusion-aware fusion operation to prevent distortion in unobserved regions. Our method achieves impressive results comparable to optimization-based methods while operating up to 500 times faster. Our framework is well-suited for applications such as semantic editing.
    \vspace{0pt}
    \blfootnote{$*$~Equal contributions; ${\dagger}$~Corresponding author.}
\end{abstract}

\section{Introduction}

Recent advances in 3D-aware generative adversarial networks (GAN) (\eg ~\cite{chan2022efficient,chan2021pi,gu2021stylenerf,or2022stylesdf}) have integrated neural radiance fields (NeRF)~\cite{mildenhall2020nerf} into style-based generation, resulting in remarkable success in generating highly realistic 3D portraits.
Similar to 2D GAN inversion techniques, 3D GAN inversion also projects images into the latent space of pre-trained 3D GAN models. 
However, in addition to pursuing faithful image reconstruction and preserving fined details, 3D GAN inversion also aims for accurate 3D geometry, which is critical for generating realistic and consistent multi-view images. This ability to ensure the correctness of 3D geometry is a crucial advantage of 3D GAN inversion over its 2D counterpart.

Some optimization-based methods, such as~\cite{ko20223d,xie2022high,yin20223d}, first invert the image to a latent code to obtain a coarse 3D shape. They then synthesize pseudo-multi-view images with this shape using a warping process. These pseudo images are used as supervision to optimize the parameters of the 3D GAN for more precise 3D geometry and better detail preservation. While these iterative optimization-based pipelines can prevent shape collapse and produce accurate results, the optimization process needs to be performed for each individual case, which is time-consuming and impractical for many applications.

Several encoder-based methods~\cite{lan2022self,li20223d} leverage a pre-trained 3D GAN to generate proxy data by randomly sampling latent codes and camera poses. They then employ these paired proxy data to train an encoder in a self-supervised fashion and further refine features in 3D GAN.
These encoder-based techniques offer higher efficiency but lower high-fidelity reconstruction ability. Moreover, due to a domain gap between the paired proxy data and real images, the encoder may invert the in-the-wild images to unsuitable latent codes. 
To address this issue, we study the properties of the latent space in 3D GAN and develop a geometry-aware encoder that predicts the latent code incorporating these properties. 
Furthermore, since the latent code can not preserve details of input images well, we explore the characteristics of feature space in 3D GAN to help our method improve the performance in reconstructing the original appearance. The latent and feature spaces are widely discussed in 2D GAN inversion~\cite{dinh2022hyperinverter,liu2022delving,tov2021designing}.

This paper presents a novel encoder-based method for inverting 3D images generated by EG3D~\cite{chan2022efficient}, a well-known and high-quality 3D GAN. Our approach is simple yet effective and is based on the discovery of a specific latent space within EG3D, which we call the \textbf{canonical latent space} $\mathcal{W}_c$. We find that latent codes belonging to this space can produce human faces with superior 3D shape and texture quality following two principles. Firstly, the input pose before the mapping network of EG3D must be a static canonical pose throughout the generation process. Secondly, the depth of the background should fall within a fixed range. Based on this, we propose using an encoder to invert the image into this canonical latent space using a proposed latent code discriminator and depth regularization. Although our encoder produces a suitable latent code that preserves the reasonable 3D shape of the input image, but the code is low dimension, making it challenging to capture texture details. To address this, similar to the 2D GAN inversion~\cite{alaluf2022hyperstyle,wang2022high}, we refine the selected features in EG3D based on the residual between the input image and reconstruction image with predicted latent code. Meanwhile, we also find there is a \textbf{canonical feature space} $\mathcal{F}_c$ in EG3D, so we design an alignment module to project the refined feature to the canonical feature space. Moreover, we introduce an occlusion-aware fusion operation to avoid texture and shape distortions in reconstructing invisible regions.

We conduct comprehensive experiments to evaluate the efficiency and effectiveness of our method, and the results demonstrate that our method achieves high-quality inversion comparable to optimization-based methods and superior to other encoder-based methods. We also demonstrate our approach is effective in 3D-aware editing applications. See Fig.~\ref{fig:teaser} for some examples of inversion and editing of human portraits and cat faces obtained by our method. Our contribution can be summarised as following:

\begin{itemize}
\vspace{-0.1cm}
\item We conduct an exploration on the latent space and feature space of EG3D and discover the presence of a canonical attribute.
\vspace{-0.2cm}
\item Based on our analysis of the characteristics of the canonical latent space, we propose a geometry-aware encoder by utilizing a canonical latent discriminator and depth regularization.
\vspace{-0.2cm}
\item We introduce an alignment module based on the canonical feature space and an occlusion-aware method to refine the selected features of the generator and supplement high-quality details.
\vspace{-0.2cm}
\item The proposed method exhibits competitive performance with existing methods in human portraits, both qualitatively and quantitatively. Additionally, our method can also generalize to the cat faces and is effective in 3D editing tasks.
\end{itemize}

\section{Related Work}

\subsection{Generative 3D-Aware Image Synthesis.}
Generative Adversarial Network~\cite{goodfellow2014generative} made significant advancements in generating both 2D photorealistic images~\cite{karras2019style, Karras2020stylegan2} and 3D-Aware consistent images.
In 3D-aware image generation, some works build 3D GAN based on explicit 3D representations, such as point-cloud~\cite{tatarchenko2016multi}, voxel~\cite{nguyen2019hologan, nguyen2020blockgan, tulsiani2017learning}, and mesh~\cite{goel2020shape, henderson2020leveraging, kanazawa2018learning}. However, these models often fail to generate high-resolution 3D scenes with high-quality details.

Since NeRF~\cite{mildenhall2020nerf} achieved outstanding multi-view image synthesis, a sequence of works has begun to use implicit 3D representation and volume rendering for 3D generation~\cite{chan2022efficient, chan2021pi, deng2022gram, gu2021stylenerf, niemeyer2021giraffe,   or2022stylesdf}.
Pi-GAN~\cite{chan2021pi} is the first attempt at using GAN to generate NeRF for 3D-aware image synthesis. After that, StyleNeRF~\cite{gu2021stylenerf} proposes a view-consistent upsampling module to generate high-resolution images. StyleSDF~\cite{or2022stylesdf} combines an SDF-based volume renderer and a 2D CNN network to obtain geometry information. GRAM~\cite{deng2022gram} introduces 2D manifolds to regulate point sampling strategy in volume rendering.
EG3D ~\cite{chan2022efficient} leverages StyleGAN2~\cite{Karras2020stylegan2}'s structure and designs a computationally efficient 3D representation called tri-plane. It can synthesize high-resolution view-consistent images and produces high-quality 3D geometry. We explore EG3D's latent space and build our inversion encoder based on EG3D.

\subsection{GAN Inversion}
GAN Inversion is an inverse process of the generation that projects the given image to corresponding latent code for faithful reconstruction from the source image. Thus, the image editing process can be conducted by semantic editing on the latent code~\cite{patashnik2021styleclip,shen2020interpreting,zhu2020domain}.
In 2D GAN, inversion methods are mainly divided into three categories. Optimization-based methods~\cite{abdal2019image2stylegan,   collins2020editing, wu2021stylespace} optimize the generator's latent code or parameters both,  achieve high reconstruction quality, but are time-consuming. 
Encoder-based methods~\cite{alaluf2021restyle,richardson2021encoding, tov2021designing} train an encoder to project input images to the latent codes. These methods can get faster inference than optimization-based methods but unsatisfactory inversion fidelity. 
Hybrid methods ~\cite{chai2021ensembling, roich2022pivotal, zhu2020domain} use an encoder to get a latent as initialization, then optimize the latent code and generator parameters, which is still time-consuming. 

Unlike 2D methods, 3D GAN Inversion should also consider multi-view-consistence. Directly transferring 2D  methods to the 3D inversion is prone to suffering geometry collapse and artifacts. To solve it, Pose Opt.~\cite{ko20223d} jointly optimizes camera pose, latent codes, and generator parameters. HFGI3D~\cite{xie2022high} is an optimization-based method that extracts mesh to create pseudo-multi-view supervision. Meanwhile, encoder methods are also gradually developing. IDE-3D ~\cite{sun2022ide} uses semantic segmentation to re-train EG3D and learns an encoder for inversion to edit images. NeRF-3DE ~\cite{li20223d} uses identity contrastive learning to train a based encoder and then trains a refining encoder for better details. E3DGE ~\cite{lan2022self} trains a base encoder to learn coarse shape and texture and supplements local details with pixel-aligned features. However, existing encoder-based methods~\cite{lan2022self,li20223d} rely on synthetic paired data and may not generalize well to real-world scenarios. None of the state-of-the-art methods can handle time efficiency and 3D inversion quality simultaneously.

\begin{figure}[t]
	\begin{center}
		\vspace{-0.2in}
		\includegraphics[width=0.95\linewidth]{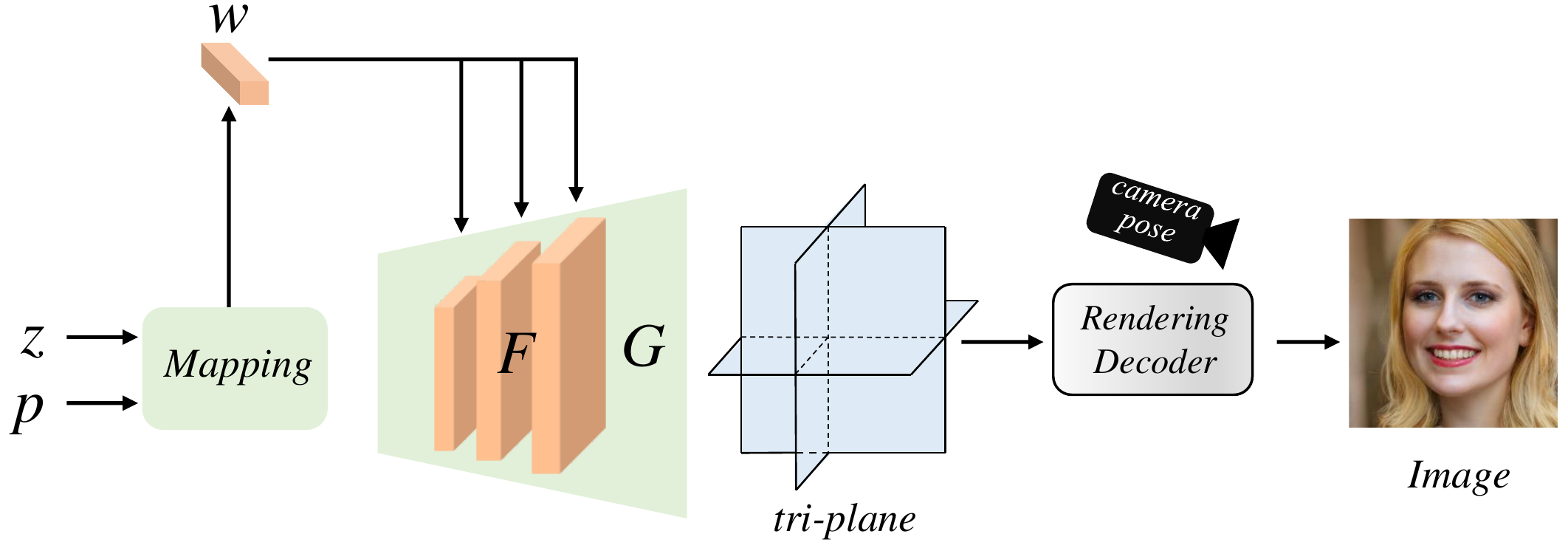}
	\end{center}
	\vspace{-0.7cm}
	\caption{\textbf{Recap of EG3D}. $z, p$ denotes a Gaussian sample and a pose condition. The relation of $w$, $F$, and tri-plane is illustrated.}
	\label{eg3d}
	
\end{figure}

\section{EG3D  Exploration}
In this section, we first briefly review the EG3D model. Then, we explore and analyze the properties of latent space and feature space in EG3D.

The pipeline of EG3D is shown in Fig~\ref{eg3d}. It first combines a Gaussian vector $z$ and a pose condition $p$ as input and builds a mapper to generate a latent code $w$, this process can be denoted as $w = mapping(z, p)$. Then this latent code $w$ will be sent to each layer of a style-based hierarchical generator to obtain the feature maps $F$. These feature maps are enlarged layer by layer. The last layer's feature map reshape to $tri\text{-}plane$, an efficient 3D representation consists of three axis-aligned feature planes ($F_{xy}$, $F_{xz}$, $F_{yz}$). With this efficient 3D representation, EG3D samples point features from the tri-plane to synthesize multi-view images or depths by a rendering decoder $R_{dec}$ with volume rendering~\cite{mildenhall2020nerf}. 
Similar to the 2D GAN, the latent code $w$ belongs to a  latent space  $\mathcal{W}$, and all feature maps in the generator construct the $\mathcal{F}$ space.

\begin{figure*}[t]
	\begin{center}
		\vspace{-0.2in}
		\includegraphics[width=1\linewidth]{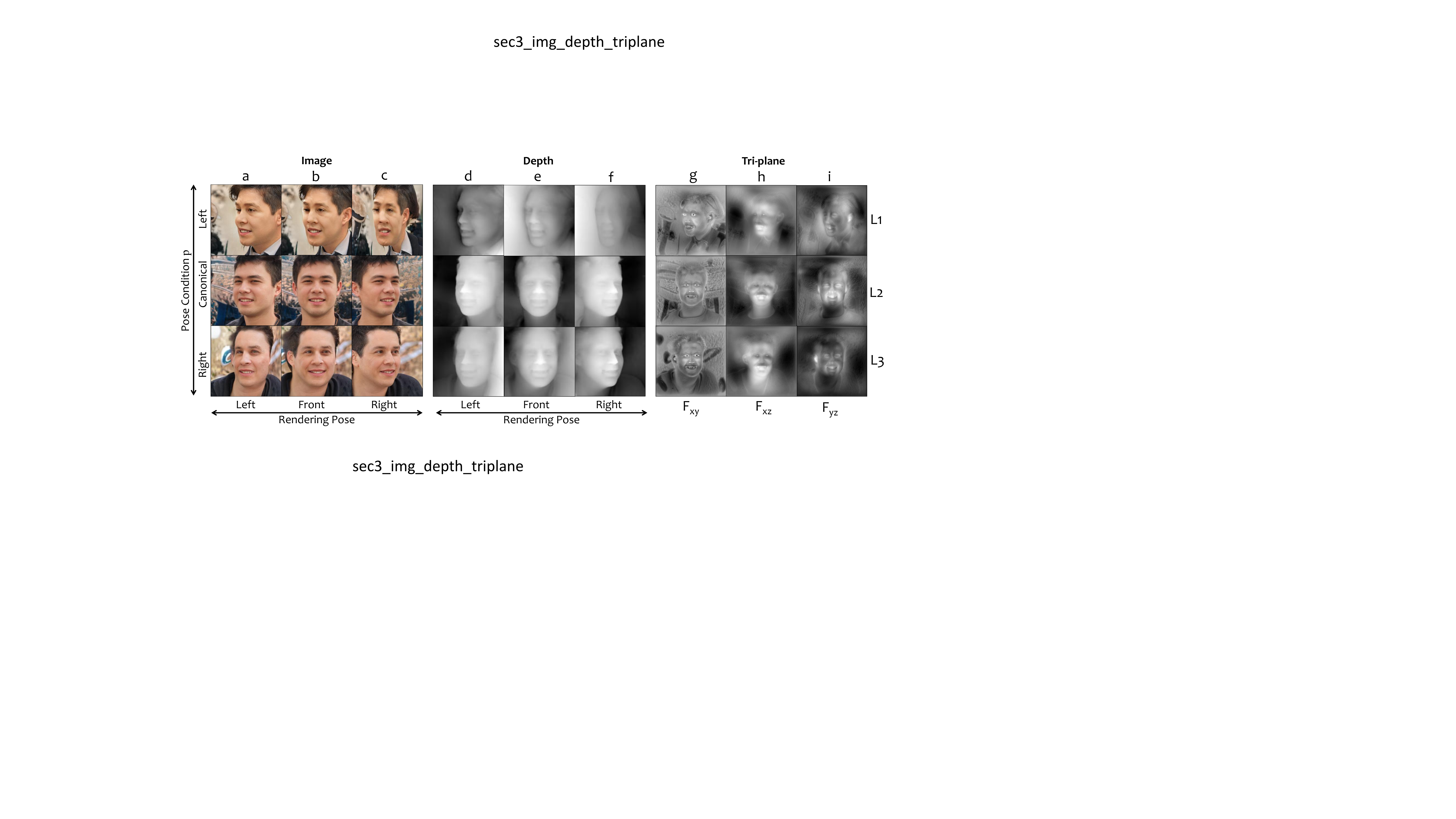}
	\end{center}
	\vspace{-0.6cm}
	\caption{\textbf{Illustration of canonical Pose condition.} We synthesize multi-view images for three 
 pose conditions $p$, including left, right, and canonical. As 
 shown in  a $\sim$ c, the canonical pose condition in row L$_2$ has no distortion face shape. Moreover,  depth maps in row L$_2$ demonstrate that images on the canonical pose condition have significant foreground and background differences. In contrast, flattened shape depth maps in rows L$_1$ and L$_3$ have foreground and background confusion. Columns g $\sim$ i reveal the tri-plane is also related to the pose condition.}
	\label{Sec3_pic_depth_triplane}
	\vspace{-0.3cm}
\end{figure*}

\subsection{Canonical Latent Space $\mathcal{W}_c$}
\label{w space}


In order to investigate the properties of the latent space $\mathcal{W}$ and its relationship with pose conditioning, we adopt a fixed $z$ and sample various pose conditions to generate corresponding latent code $w$. Fig.~\ref{Sec3_pic_depth_triplane} illustrates that when $w$ is conditioned on an unsuitable pose $p$ far away from the canonical pose, the generated 3D shapes fail to generalize to other views synthesis. Only when the pose condition is canonical can the synthesized images maintain shape and texture consistency across multiple views. Meanwhile, we find a clear depth gap between the face and background regions when generating good results with the canonical pose condition. Under canonical pose, the depth of the background is restricted to a fixed range, while the face and background regions are confused for the other pose conditions.

These observations lead us to identify a specific subspace of  $\mathcal{W}$  intuitively, which we call it \textbf{canonical latent space}, denoted as $\mathcal{W}_c$,  that is responsible for ensuring reasonable 3D shape and texture across multiple views. There are two principles of this canonical latent space. Firstly, the input pose before the mapping network of EG3D must be a static canonical pose. Secondly,  the depth of the background should fall within a fixed range to distinguish the face and background regions. Based on these findings, we propose that a high-quality 3D-GAN inversion method should first retrieve a $w$ latent code in $\mathcal{W}_c$ to ensure the best possible image synthesis results.

\subsection{Canonical Feature Space $\mathcal{F}_c$}
\label{F space}
To analyze the characteristics of $\mathcal{F}$ space, we generate the latent codes with different pose conditions to synthesize multiple tri-planes. We visualize the tri-plane as illustrated in Fig~\ref{Sec3_pic_depth_triplane}. The tri-plane also represents a canonical view when the pose condition is canonical (\ie., the columns g $\sim$ i in row L$_2$). 
It indicates that there is also a specific subspace of $\mathcal{F}$, we call it \textbf{canonical feature space}, denoted as $\mathcal{F}_c$, and the canonical latent space $\mathcal{W}_c$ produces the canonical feature space $\mathcal{F}_c$ with the generator layers. The $\mathcal{F}_c$ can supply sufficient facial information for rendering multi-view images.

\section{Method}
\begin{figure*}[ht]
	\begin{center}
		\vspace{-0.2in}
		\includegraphics[width=0.85\linewidth]{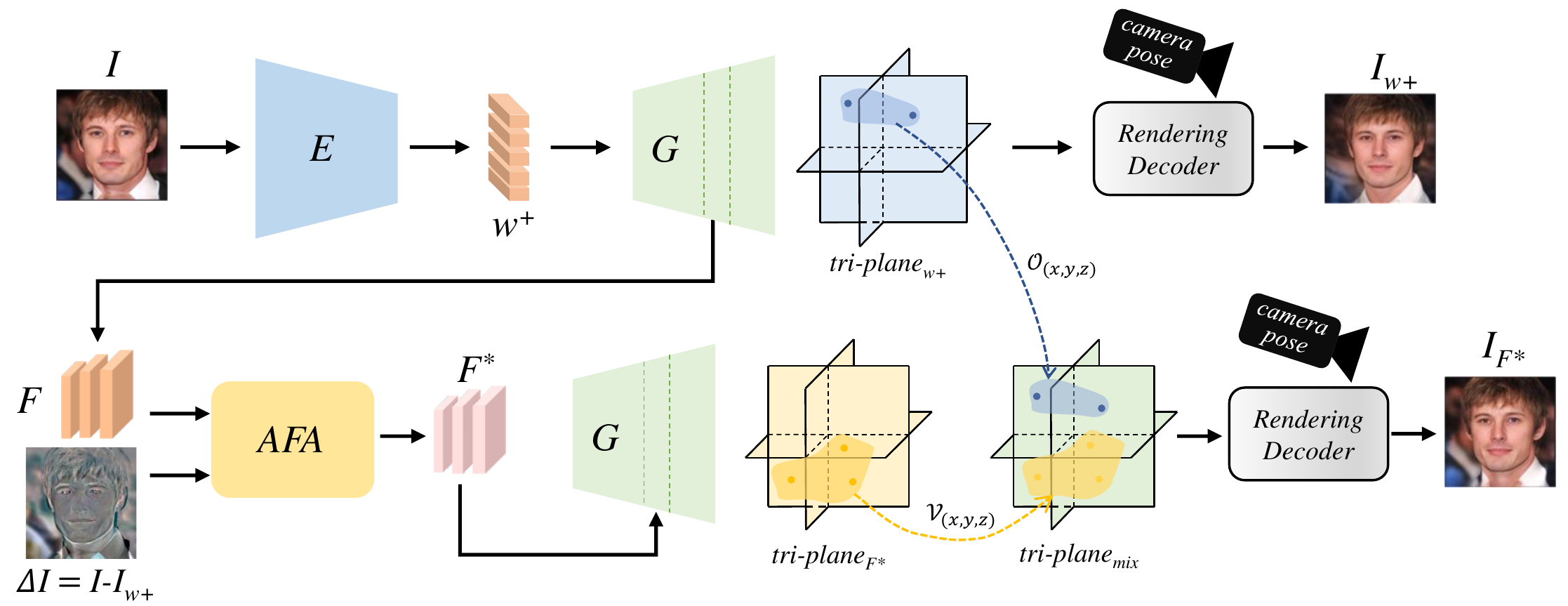}
	\end{center}
	\vspace{-0.7cm}
	\caption{\textbf{Overview of our method}. Our framework could be divided into two parts. (1) $\mathcal{W}$ space inversion. We design an encoder $E$ to invert input image $I$ into $w^+$ latent codes. The $w^+$ latent codes are fed into a pre-trained EG3D generator $G$ to get $tri$-$plane_{w^+}$ and rendered into reconstruction image $I_{w^+}$. (2) Complement the $\mathcal{F}$ space. We calculate the image residual $\Delta I$ between the input image and its reconstruction and propose AFA module to refine the F latent maps. The modified latent maps $F^*$ are transformed into $tri$-$plane_{mix}$ by occlusion-aware mix and rendered into the fine detailed inversion image $I_{F^*}$.}
	\label{fig:framework}
	\vspace{-0.4cm}
\end{figure*}

The pipeline of our method is shown in Fig.~\ref{fig:framework}. We first project the input image $I$ to the latent code $w^+ \subset  \mathbb{R}^{14 \times 512}$ with our encoder, where $w^+ = (w_0,w_1,...,w_{13})$ is an extension of latent code $w$ and can preserve the details better as mentioned in  2D GAN inversion ~\cite{richardson2021encoding, tov2021designing}. Then we calculate the residual between the $I$ and the generated image $I_{w^+}$ based on $w^+$ to refine the features of the generator. In the following Sec.~\ref{Encoder} and Sec.~\ref{Feature}, we will describe how we ensure the latent code $w^+$ and refined feature belong to the canonical space, respectively. Furthermore, we introduce an occlusion-aware mix tri-plane in Sec.~\ref{Occlusion} to avoid the distortion of invisible regions reconstruction. In Sec.~\ref{editing}, we exhibit how to perform 3D semantic editing with our inversion method.

\subsection{Geometry-aware Encoder}
\label{Encoder}
We design a geometry-aware encoder to obtain the latent code $w^+$ given the input image $I$ and corresponding rendering camera pose $c$. We generate the inverted latent code $w^+$ according to the backbone's pyramid feature progressively. It is worth noting that we choose Swin-transformer~\cite{liu2021swin} as the backbone and further add attention modules at different scale feature layers for different level latent codes (more details of encoder architecture are in supplementary materials). After projecting the input image to the latent code $w^+$, we can synthesize the $ tri\text{-}plane_{w^+}$ through EG3D generator $G$ and use rendering decoder $R_{dec}$ to get the reconstruction image $I_{w^+}$ under the rendering camera pose $c$.
\begin{equation}
    \begin{footnotesize}
    \begin{aligned}
        w^+ &= E(I), \\
        I_{w^+}&= R_{dec}(G(w^+), c)
    \end{aligned}
    \end{footnotesize}
\end{equation}

To make the inverted $w^+$ not deviate far from the canonical space $\mathcal{W}_c$ discussed in Sec.~\ref{w space}, we use a canonical latent discriminator~\cite{nitzan2020face} to discriminate between $w_c \in \mathcal{W}_c$ and $w^+$. The canonical latent discriminator, denoted as $\mathcal{D}_{\mathcal{W}_c}$, is trained in an adversarial scheme with the non-saturating GAN loss \cite{goodfellow2014generative} and $R_1$ regularization \cite{mescheder2018training}, the loss of adversarial learning process can be written as below:
\begin{equation}
\begin{gathered}
\mathcal{L}^{\mathcal{D}_{\mathcal{W}_c}}_{\text{adv}} =  -{\mathbb{E}}[\log \mathcal{D}_{\mathcal{W}_c}(w_c)]-{\mathbb{E}}[\log (1-\mathcal{D}_{\mathcal{W}_c}(w_i))] \\  + 
\frac{\gamma}{2} {\mathbb{E}}\left[\left\|\nabla \mathcal{D}_{\mathcal{W}_c}(w_c)\right\|_2\right],
\end{gathered}
\end{equation}
\begin{equation}
    \mathcal{L}_{\text {adv}}^E=-{\mathbb{E}}\left[\log {\mathcal{D}}_{\mathcal{W}_c}(w_i)\right],    
\end{equation}
where $\gamma$ is the weight of $R_1$ regularization, the 
$ \mathcal{L}^{{\mathcal{D}}_{\mathcal{W}_c}}_{\text{adv}} $ and $\mathcal{L}_{\text {adv}}^E$ are the loss for the latent discriminator $\mathcal{D}_{\mathcal{W}_c}$ and encoder respectively. $\mathcal{D}_{\mathcal{W}_c}$ operates on $w_c$ and each entry $w_i$ of $w^+$.

As discussed in Sec.~\ref{w space}, the background depth of an integrated face should be limited in a range. To estimate the background depth, we first random sample 0.1 million front images with the canonical pose condition, then we obtain the depth maps and background regions of these images.  We use the parsing method~\cite{parsenet} to get a binary mask $M$, where $M=1$ indicates the background region and 0 otherwise.  Finally, we calculate the average depth $D_{avg}$ of background regions of these samples, and we set the $D_{avg}$ as a regularization to constrain the background depth during inversion, the background regularization can be written as below:

\begin{equation}
    \mathcal{L}_{BG} =  \left \| D \odot M - D_{avg} \odot M \right \|_2.
\end{equation}
Here, $D$ and $M$ are the depth map and background mask inverted image under the front-view pose.

In addition, we use reconstruction loss function $\mathcal{L}_{rec}$ follow
pSp~\cite{richardson2021encoding} in the training stage. The $\mathcal{L}_{rec}$ is composited by pixel-wise $\mathcal{L}_2$ loss, perceptual loss $\mathcal{L}_{LPIPS}$~\cite{zhang2018unreasonable}, and facial identity similarity loss $\mathcal{L}_{ID}$ with pre-trained ArcFace network~\cite{deng2019arcface}:
\begin{equation}
    \begin{footnotesize}
    \begin{aligned}
        \mathcal{L}_{rec} &= \left \| I_{w^+} -  I \right \|_2  + \mathcal{L}_{LPIPS}(I_{w^+}, I) + \mathcal{L}_{ID}(I_{w^+},I).
    \end{aligned}
    \end{footnotesize}
\end{equation}

In summary, the total loss for training our encoder can be written as:
\begin{equation}
    \mathcal{L}_{total} = \mathcal{L}^\mathcal{D}_{\text{adv}} + \mathcal{L}_{\text{adv}}^E   + \mathcal{L}_{BG} + \mathcal{L}_{rec} .
\end{equation}

\subsection{Adaptive Feature Alignment}
\label{Feature}
It is non-trivial to restore fine-detailed texture with only a highly compressed $w$ latent code. Thus, adjusting high-dimension feature maps in the $\mathcal{F}$ space is natural to achieve high-fidelity inversion as mentioned in related 2D GAN inversion methods ~\cite{alaluf2022hyperstyle, dinh2022hyperinverter,wang2022high}. Moreover, it is worth noting that the feature maps $F$ generated with our inverted $w^+$ are in the canonical feature space $\mathcal{F}_c$.

We calculate the residual $\Delta I$ between the  $I$ and $I_{w^+}$ and hope to refine the feature maps to eliminate this residual. However, directly setting the residual as input and predicting the refined feature maps 
 with CNN to modify $F$ is invalid. This is because the residual is highly entangled with the view of the input image, while $F$ is in $\mathcal{F}_c$. 
 When the view is not canonical, the refined feature maps with this residual can not project to the $\mathcal{F}_c$ space simply. Thus, aligning the residual feature maps to the $\mathcal{F}_c$ space is necessary to synthesize high-quality images, as mentioned in Sec.~\ref{F space}.

\begin{figure}
	\begin{center}
		\includegraphics[width=1\linewidth]{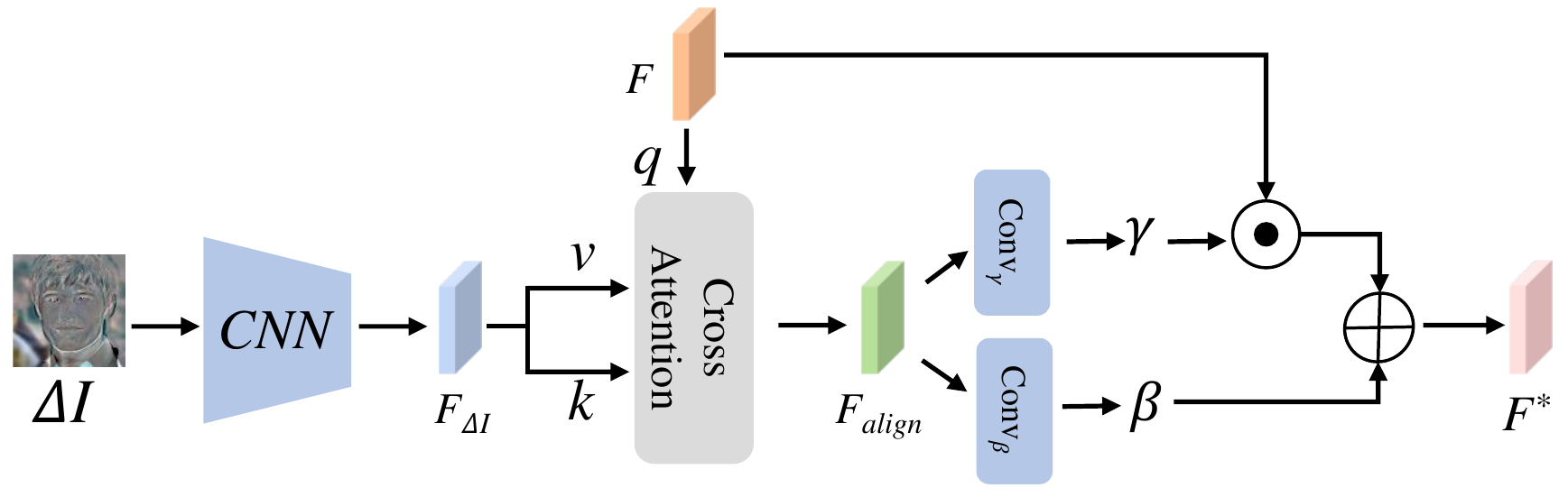}
	\end{center}
    \vspace{-0.7cm}
	\caption{\textbf{The Architecture of AFA Module}. We use a CNN to capture the feature $F_{\Delta I}$  of residual image $ {\Delta I}$, and align the $F_{\Delta I}$ to $F$ with cross attention module. Finally, we refine the $F$ with the aligned feature $F_{align}$.  }
	\label{afa}
  \vspace{-0.4cm}
\end{figure}

According to the above analysis, we propose an adaptive feature alignment (AFA) module that enables automatic canonical space alignment by cross-attention mechanism, as shown in Fig~\ref{afa}. First, we use a CNN to capture the feature $F_{\Delta I}$ of $\Delta I$, and we take $F_{\Delta I}$  as the key $K $, value $V $, and the feature map $F$ as a query $Q$. Then, we utilize cross-attention modules to align the $F_{\Delta I}$ with $F$ and output a feature $F_{align}$. Since the $F$ is already in the canonical feature space $\mathcal{F}_c$, and the feature $F_{align}$ can also have the properties of $\mathcal{F}_c$ space. Finally, we use the $F_{align}$ to predict the scale map $\gamma$ and translation map $\beta$ to refine the $F$ with Feature-wise Linear Modulation (FiLM) ~\cite{perez2018film}. The whole process of our AFA module can be formulated as follows: 
\begin{equation}\label{eq:channel dimension}
\begin{aligned}
   & F_{\Delta I} = CNN(\Delta I), \\
   & Q, K, V= W_Q F, W_K F_{\Delta I}, W_V F_{\Delta I},\\
   & F_{align} = Attention(Q ,K ,V ), \\
   & \gamma, \beta = Conv_{\gamma}(F_{align}), Conv_{\beta}(F_{align}), \\
   & F^* = \gamma  \odot F + \beta ,\\ 
\end{aligned}
\end{equation}
where $W_Q , W_K , W_V$ are learnable projection heads and $d$ is the dimension of feature, the $F^*$ is the final output of our AFA module. 
With this adaptive alignment, we achieve a detailed complement of $\mathcal{F}_c$ space. Note that we adopt $\mathcal{L}_{total}$ and a $\mathcal{L}_2$ regularization loss on $\Delta F = F^*-F$ when training the AFA module.

\subsection{Occlusion-aware Mix Tri-plane}
\label{Occlusion}

The image residual $\Delta I$ only contains information in the visible area of the input view. Using single-view residuals to modify the entire $F$ will lead to distortion at the occlusion region. To solve this problem, we propose the Occlusion-aware Mix Tri-plane.

The tri-plane contains all points in the 3D space which are divided into visible points set and occlusion points set according to the input view. Here, we define the 3D points viewed in the input image in tri-plane as the visible points set $\mathcal{V}_{(x,y,z)}$, points in the occlusion regions as occlusion points set $\mathcal{O}_{(x,y,z)}$. We can get the depth map $D$ under the input view, and the two points set can be defined by: 
\begin{small}
\begin{equation}
\begin{aligned}
    \mathcal{V}_{(x,y,z)}=\{[x, y, z]^T : &[x, y, z]^T = \pi D(u,v)\mathcal{K}^{-1}[u,v,1]^T \}, \\
    \mathcal{O}_{(x,y,z)} &= \mathbb{R}^3 \setminus \mathcal{V}_{(x,y,z)} ,    
\end{aligned}
\end{equation}
\end{small}
where $[u,v,1]$ is the pixel homogeneous coordinate of the input image, $\pi$ is the extrinsic camera parameter matrix of the input view, $D(u,v)$ is the depth of pixel $[u,v]$, $\mathcal{K}$ is the intrinsic camera parameter matrix. 

After determining the visible area, we can supplement the refined details of the input view to the tri-plane.
We define the occlusion-aware mix tri-plane as follows:
\begin{equation}
\begin{aligned}
 tri\text{-}plane_{w^+} &= G(w^+),\\
 tri\text{-}plane_{F^*} &= G(F^*), \\
tri\text{-}plane_{mix} &= tri\text{-}plane_{F^*}|_{(x,y,z) \in \mathcal{V}} \\
&+ tri\text{-}plane_{w^+}|_{(x,y,z) \in \mathcal{O}}, \\    
\end{aligned}
\end{equation}
where $w^+$ is the latent code learned from the encoder, and $F^*$ is the refined feature map. Finally, we use $tri\text{-}plane_{mix}$ with rendering decoder to synthesize images under different views when inference.

\subsection{Editing}
\label{editing}
To achieve 3D-consistent editing, we need to modify both $w^+$ and $F^*$. We first get the edited latent code $\hat{w}^+$ through classic latent code editing methods ~\cite{patashnik2021styleclip,shen2020interpreting}. Then for $F^*$, we follow FS~\cite{xuyao2022} to extract the features $F^{\hat{w}^+}$ and $F^w$ from the generator convolution layer with input $\hat{w}^+$ and $w$, respectively. Finally, we calculate the difference between these two features and add it to the $F^*$ to get the edited features $\hat{F^*}$:
\begin{equation}
\begin{aligned}
    \hat{F^*} = F^* + F^{\hat{w}^+} - F^w.
\end{aligned}
\end{equation}
With the modified $\hat{w}^+$ and $\hat{F^*}$, we use the method in section \ref{Occlusion} to get edited mix tri-plane, and use rendering decoder to synthesize 3D-consistent edited images.

\begin{figure*}
	\begin{center}
		\includegraphics[width=0.9\linewidth]{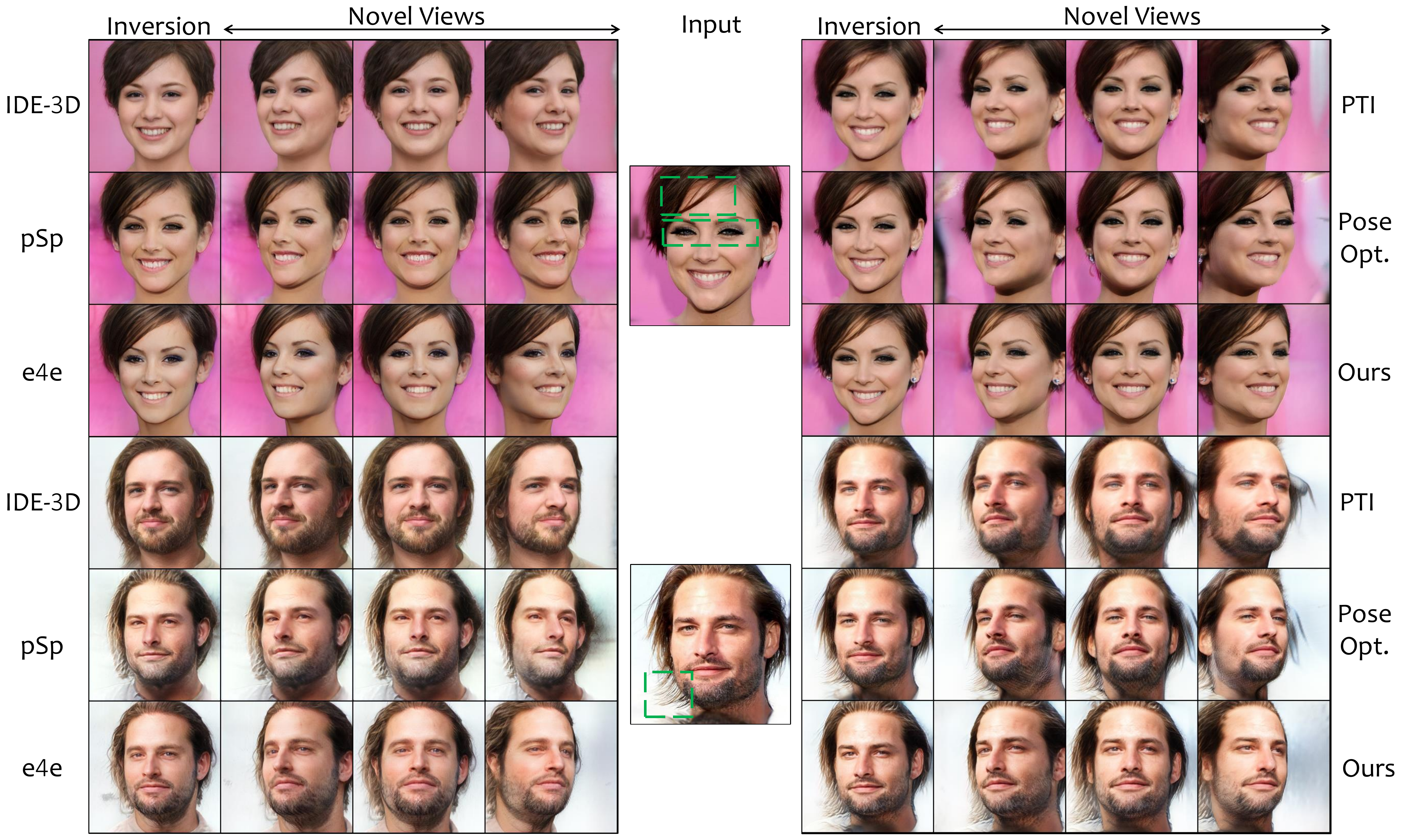}
	\end{center}
	\vspace{-0.5cm}
	\caption{\textbf{Visual comparison of inversion and novel-view synthesized results between our method and others.} Our method surpasses all the encoder-based methods and is competitive with optimization methods in recovering texture details, \ie our method prevents the eye orientation from changing with the camera view and artifacts in the novel view. }
	\label{Qualitative Comparision}
\end{figure*}

\section{Experiments}
\noindent \textbf{Experiment setup.} We evaluate our method on human faces and cat faces. For human faces, we use FFHQ ~\cite{abdal2019image2stylegan} dataset for training, and CelebA-HQ~\cite{karras2018progressive} and MEAD ~\cite{wang2020mead} for evaluation. For the cat faces, we use AFHQ-CAT ~\cite{choi2020stargan} for training and evaluation. In addition, we employ the pre-trained EG3D generator and obtain the corresponding camera parameters of each image following the official implementation in EG3D.


\noindent \textbf{Baselines.}
We compare our method with encoder (IDE-3D~\cite{sun2022ide}) and optimization-based (PTI~\cite{roich2022pivotal} and Pose Opt.~\cite{ko20223d}) baselines in these two aspects. 
Moreover, we build two encoder baselines imitating the state-of-the-art StyleGAN inverters, pSp~\cite{richardson2021encoding} and e4e~\cite{tov2021designing}, and present the model structure in the supplementary.


\subsection{Qualitative analysis}

\noindent \textbf{Source View Inversion Results.} 
Fig.~\ref{Qualitative Comparision} shows the inversion performance comparison between ours and other methods. Our method's hair details and facial similarity are obviously competitive with optimization methods and surpass all the encoder-based methods.

\noindent \textbf{Novel View Synthesis Results.}
The comparison results are shown in  Fig.~\ref{Qualitative Comparision}. 
Inspecting the eye gaze direction and person's identity preservation when changing to novel views can reveal the differences. It can be seen that on some extreme side-face camera poses, some optimization methods produce artifacts and geometry distortions. 
Our method can keep synthesizing 3D-consistent images faithful to the person's identity in the input.

\begin{table*}[h] 
    \Huge
    \renewcommand\arraystretch{1.1}
    \centering  
    \caption{ \textbf{Comparison with state-of-the-art methods on inversion performance.} The numbers in bold are the best performance of the optimization and encoder-based methods, respectively. Metrics of Reconstruction Quality are calculated under the input view. Metrics of Novel View Quality are the average of multi-view. The yaw angle is the source view angle. } 
    \label{Input view Inversion metrics}  
    \setlength{\tabcolsep}{7 pt}
    \resizebox{1\linewidth}{!}{
    \begin{tabular}{l|c|cccccc|cccccccc}
    \toprule 
     & \multicolumn{1}{l|}{} & \multicolumn{6}{c|}{Reconstruction Quality} & \multicolumn{8}{c}{Novel View Quality} \\ \cline{3-16} 
    Category & \multicolumn{1}{l|}{Method} & \multicolumn{6}{c|}{Input View} & \multicolumn{4}{c|}{yaw=$\pm 60^{\circ}$} & \multicolumn{4}{c}{yaw=$0^{\circ}$} \\ \cline{3-16} 
     & \multicolumn{1}{l|}{} & MSE $\downarrow$ & LPIPS $\downarrow$ & FID $\downarrow$ & ID $\uparrow$ & \multicolumn{1}{l}{Geo. Err.} $\downarrow$ & \multicolumn{1}{l|}{ Time (s)$\downarrow$}  & \multicolumn{1}{l}{LPIPS} $\downarrow$ & \multicolumn{1}{l}{FID} $\downarrow$ & \multicolumn{1}{l}{ID} $\uparrow$ & \multicolumn{1}{l|}{Geo. Err. $\downarrow$}  & \multicolumn{1}{l}{LPIPS} $\downarrow$ & \multicolumn{1}{l}{FID} $\downarrow$ & \multicolumn{1}{l}{ID} $\uparrow$ & \multicolumn{1}{l}{Geo. Err. $\downarrow$}  \\ \hline
    \multirow{2}{*}{Optimization} & PTI~\cite{roich2022pivotal} & \textbf{0.0163} & 0.0918 & 18.0 & 0.8417 & 0.0850 & \textbf{115.2} & 0.2926 & 69.8 & 0.621 & \multicolumn{1}{c|}{0.1481} & \textbf{0.1925} & 53.5 & 0.718 & 0.1006 \\
     & Pose Opt.~\cite{ko20223d} & 0.0164 & \textbf{0.0879} & \textbf{17.5} & \textbf{0.8390} & \textbf{0.0759} & 219.7 & 0.4477 & 132 & 0.321 & \multicolumn{1}{c|}{0.1763} & 0.2436 & 60.2 & 0.605 & 0.1504 \\ \hline
    \multirow{5}{*}{Encoder-based} & IDE-3D~\cite{sun2022ide} & 0.1056 & 0.2806 & 43.6 & 0.7194 & 0.1835 & \textbf{0.0659} & 0.3317 & 79.0 & 0.505 & \multicolumn{1}{c|}{0.2914} & 0.3074 & 63.7 & 0.513 & 0.2528 \\
     & pSp-3D~\cite{richardson2021encoding} & 0.0726 & 0.2082 & 31.2 & 0.7817 & 0.1846 &  0.1284& 0.3223 & 65.7 & 0.591 & \multicolumn{1}{c|}{0.1287} & 0.3132 & 65.2 & 0.671 & 0.1789 \\
     & e4e-3D~\cite{tov2021designing} & 0.0808 & 0.2281 & 33.0 & 0.7359 & 0.1161 & 0.1303 & 0.2495 & 57.7 & 0.588 & \multicolumn{1}{c|}{0.1269} & 0.2394 & 53.8 & 0.653 & 0.1305 \\
     & $\text{Ours-}{w^+}$ & 0.0697 & 0.1981 & 30.9 & 0.7873 & 0.1100 &  0.0963& 0.2303 & 52.2 & 0.655 & \multicolumn{1}{c|}{\textbf{0.0943}} & 0.2372 & 54.1 & 0.685 & 0.0955 \\
     & Ours & \textbf{0.0301} & \textbf{0.1257} & \textbf{20.5} & \textbf{0.8752} & \textbf{0.0987} & 0.2262 & \textbf{0.2192} & \textbf{50.9} & \textbf{0.700} & \multicolumn{1}{c|}{0.0950} & 0.2204 & \textbf{51.7} & \textbf{0.743} & \textbf{0.0944} \\
     \bottomrule
\end{tabular}}
\end{table*}

\subsection{Quantitative analysis}
 We calculate the mean square error (MSE), LPIPS \cite{zhang2018unreasonable}, and FID \cite{heusel2017gans} between the source image and its inversion. We adopt a pre-trained face key point detector parsenet \cite{parsenet} to calculate the identity similarity. To measure the geometry properties, we followed \cite{chan2022efficient,shi2021lifting}, utilizing a pre-trained 3D face reconstruction model to extract a “pseudo” ground truth depth map from the source image. Then, we obtain the depth of inversion and normalize all depth maps to zero means and unit variance to calculate geometry error (Geo. Err.) by using average $\mathcal{L}_2$ distance.

\noindent \textbf{Source View Inversion Results.} Table ~\ref{Input view Inversion metrics} presents the quantitative comparison of the input view image reconstruction on CelebA-HQ test dataset. Our method is two orders of magnitude faster than the optimized methods with similar performance, which is notably better than other encoder-based methods.

\noindent \textbf{Novel View Synthesis Results.} In order to measure the multi-view consistency and the synthesized quality, we evaluate our method and baselines with a multi-view dataset, MEAD ~\cite{wang2020mead}. Each person in MEAD  has five images with increasing yaw angles. We select one angle as the input view, invert the input image and synthesize the other four novel views. Then we use the same metrics to calculate the difference between the synthesized images and the ground truth. The quantitative comparison between the novel view synthesized results and the ground truth in different views is shown in Table~\ref{Input view Inversion metrics}. Our method shows similar performance with PTI when the input image is a front face (yaw = $0^{\circ}$). When inverting side faces, the difficulty of inversion increases drastically when the yaw angle increases. We calculated the metrics of the largest yaw angle in MEAD (yaw = $\pm 60^{\circ}$), and our model has robust performance when the test yaw angle varies. However, the performance of optimization methods degrades remarkably at an extreme angle.

\subsection{Ablation Analysis}
We present ablation studies with qualitative analysis of our proposed method below, and the corresponding numerical ablation results are in our supplementary.

\begin{figure}
	\begin{center}
		\includegraphics[width=1\linewidth]{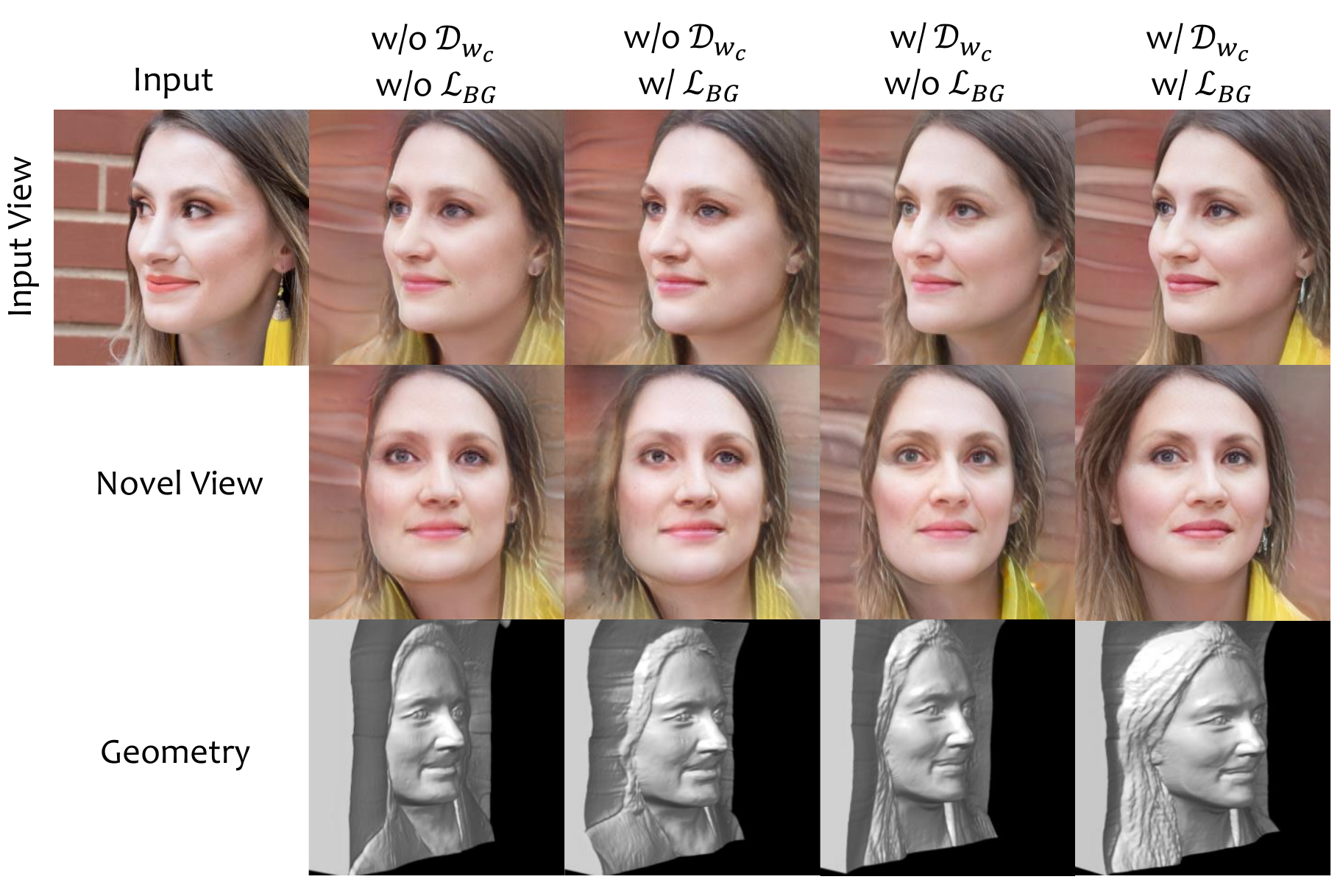}
	\end{center}
    \vspace{-0.5cm}
	\caption{\textbf{Ablation of Geometry-aware Encoder.} All images are synthesized only with $w^+$. Without both modules, $w^+$ fails to generate integrated 3D geometry and texture. The $L_{BG}$ helps distinguish foreground and background. With $\mathcal{D}_{\mathcal{W}_c}$, we can get good inversion under the input view but fail under the novel view. With both modules, $w^+$ can obtain an integrated side face of the woman, which is occlusion under the input view.}
	\label{ablation_encoder}
 \vspace{-0.5cm}
\end{figure}

\noindent \textbf{Geometry-aware Encoder.}
We conduct ablation experiments and make qualitative comparisons to verify that our proposed canonical latent discriminator ($\mathcal{D}_{\mathcal{W}_c}$) and background regularization ($\mathcal{L}_{BG}$) contribute to restoring 3D geometry. Generally, recovering the geometry from a side face input is more challenging and we test such a case. As shown in Fig.~\ref{ablation_encoder}, only if we adopt the canonical discriminator and the background regularization could we obtain a satisfactory foreground and background geometry.

\begin{figure}[!t]
	\begin{center}
		\includegraphics[width=\linewidth]{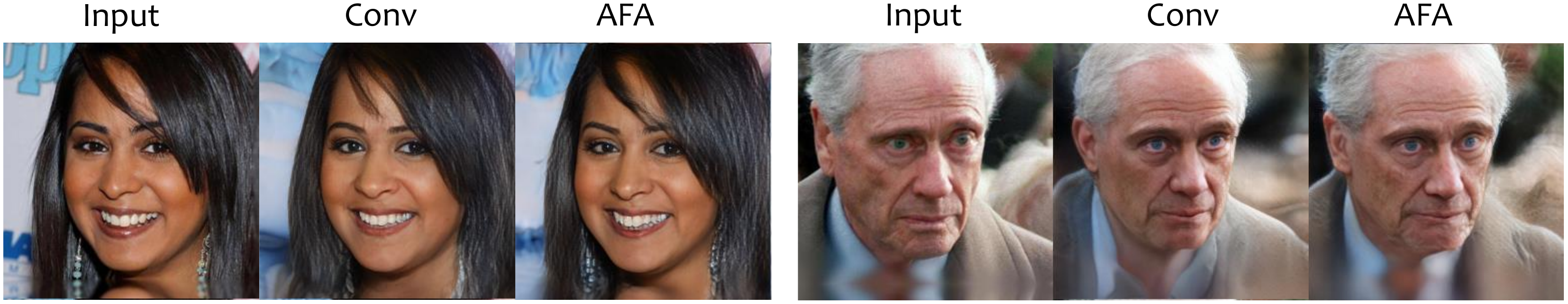}
	\end{center}
    \vspace{-0.5cm}
	\caption{\textbf{Ablation of Adaptive Feature Alignment.} We replace AFA with a CNN network and obtain the results. Without feature alignment, it is difficult to reconstruct the hair and facial details. }
	\label{ablation_afa}
\end{figure}

\noindent \textbf{Adaptive Feature Alignment.}
We proposed Adaptive Feature Alignment (AFA) to refine the $F$ latent map and add texture details from the source view. Fig.~\ref{ablation_afa} demonstrates that we can recover more details with AFA.

\begin{figure}[!t]
	\begin{center}
		\includegraphics[width=1\linewidth]{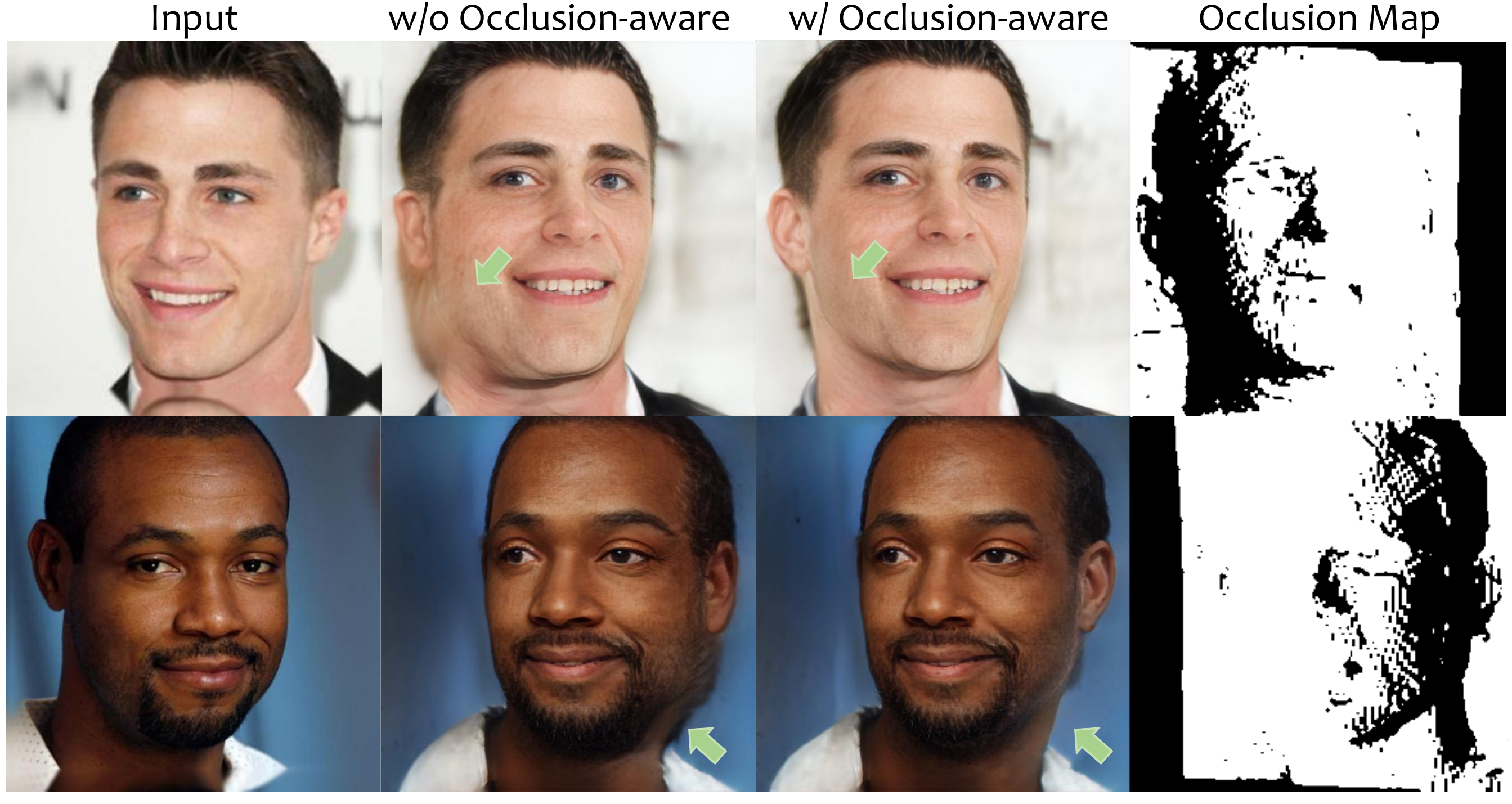}
	\end{center}
    \vspace{-0.5cm}
	\caption{\textbf{Ablation of Occlusion-aware mix tri-plane.} We visualize the occlusion regions $\mathcal{O}_{(x,y,z)}$ of the input view in the last column (black points). The green arrow points out $\mathcal{O}_{(x,y,z)}$ in the novel view, and it is obvious that the occlusion-aware mix tri-plane can avoid artifacts.}
	\label{mix_triplane}
\end{figure}

\noindent \textbf{Occlusion-aware Mix Tri-plane.} To verify the necessity of our proposed occlusion-aware mix tri-plane, we ablate this operation and visualize the novel view results in Fig.~\ref{mix_triplane}. The occlusion map shows the spatial position of surface points in ray sampling, the visible region in the original view is white, and the occlusion region is black.
Without the mix tri-plane, the occlusion area from the source view is easy to produce artifacts in novel views.

\subsection{Image Editing Results}

\begin{figure}
	\begin{center}
		\includegraphics[width=1\linewidth]{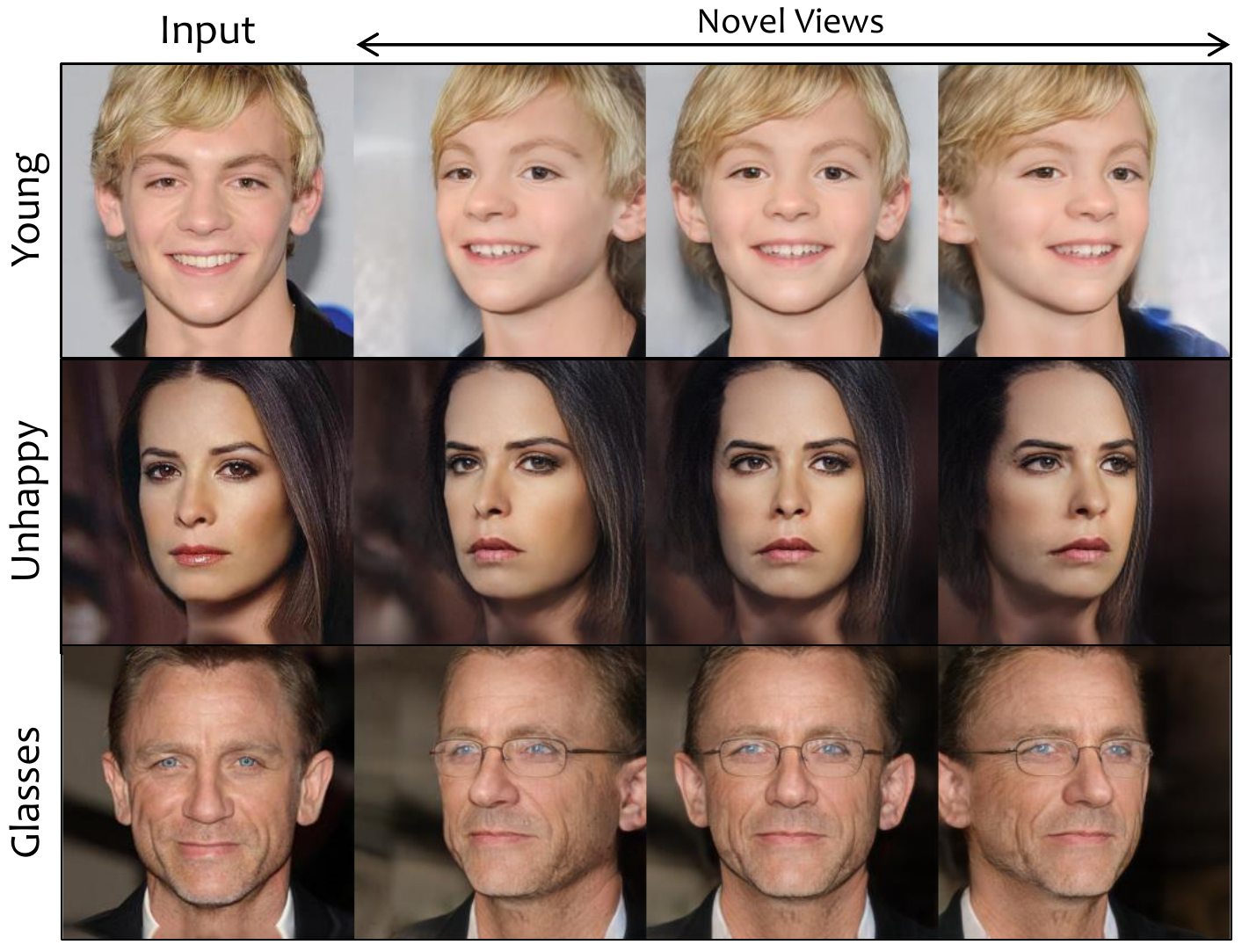}
	\end{center}
    \vspace{-0.5cm}
	\caption{\textbf{Editing result under novel views.} We add three different attributes to three faces. Our method generates both realistic editing and 3D-consistent novel views. }
	\label{edit_pic}
\end{figure}

In GAN inversion-based image manipulation, the editability and the reconstruction trade-off widely exists in both 2D and 3D scenarios. We validate the editing method mentioned in Sec.~\ref{editing} and display the multi-view edited results in Fig.~\ref{edit_pic}. We choose three different edit directions, 'Young' to reduce person's age, 'Unhappy' to change emotion, and 'Glasses' to add accessories. It is noticeable that our method not only restores the geometric and texture information of the input image but also ensures the editing consistency of multi-view. The edited results demonstrate that our inversion method can be well applied to the facial image editing tasks.  

\section{Conclusion}
 In this work, we explore the intrinsic property of EG3D's latent space and find that the key to achieve high-quality shape and texture inversion is to invert input images to the canonical space. In light of this, we design an efficient geometry-aware encoder with an adaptive feature alignment module and demonstrate their effectiveness. Besides, we propose an occlusion-aware mix tri-plane to avoid invisible region distortion in novel views. Extensive experiments demonstrate that our method has competitive inversion performance with optimization-based methods, but operates two orders of magnitude faster. In addition, our method is demonstrated to be a powerful tool for realistic and 3D view-consistent editing of facial images. 
 

{\small
\bibliographystyle{ieee_fullname}
\bibliography{arxiv}
}

\appendix
\clearpage
\newpage 
\section*{Appendix}

\begin{figure*}[t]
	\begin{center}
		\vspace{-0.2in}
		\includegraphics[width=.85\linewidth]{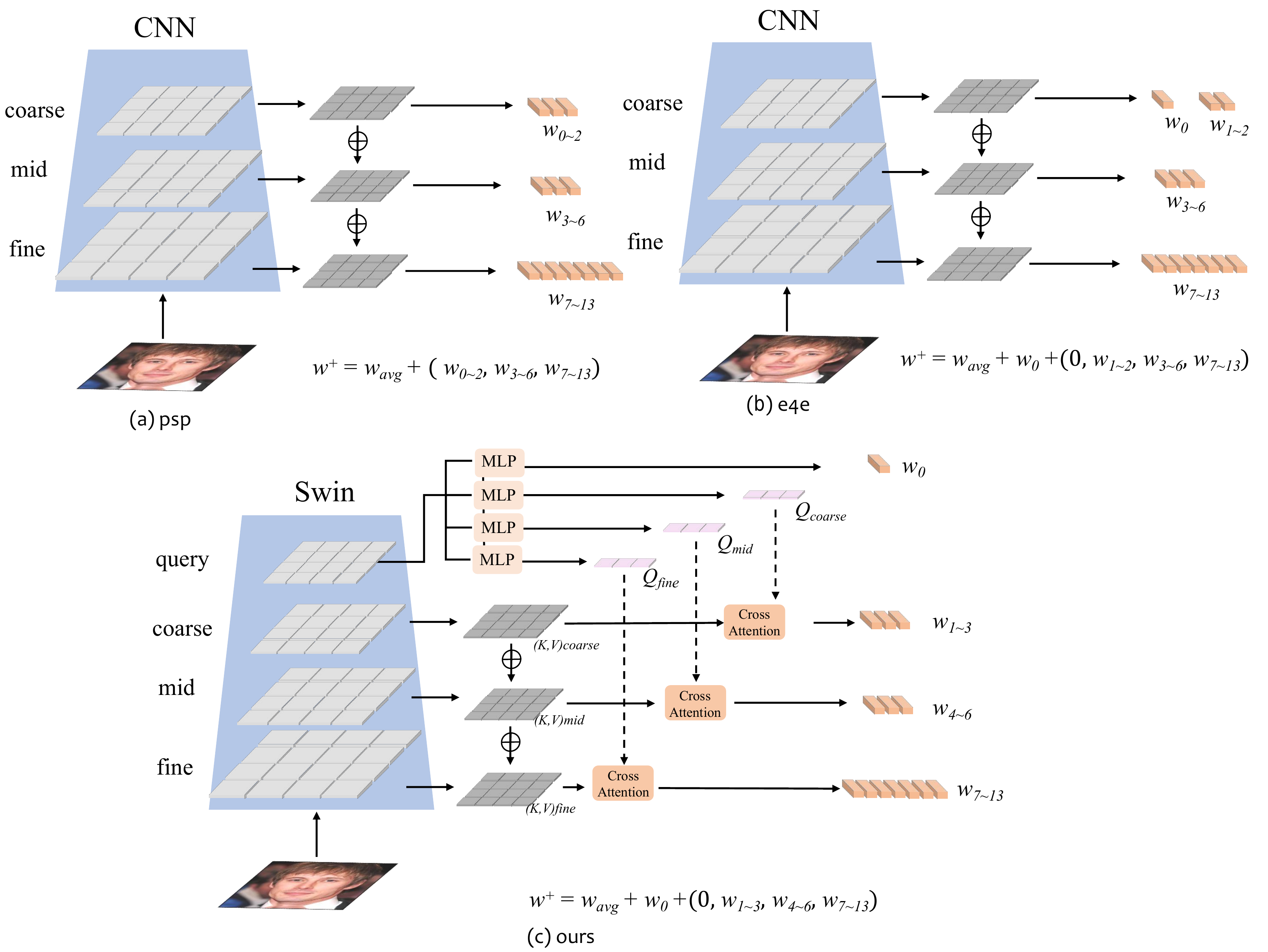}
	\end{center}
	\vspace{-0.6cm}
	\caption{\textbf{Encoder architecture comparison.} We refer to pSp~\cite{richardson2021encoding}, e4e~\cite{tov2021designing}'s model structure, refine and build up EG3D inverters respectively (see in (a),(b)). Our designed Swin-transformer based encoder is shown in (c).   }
	\label{Eswin}
 \vspace{-0.3cm}
\end{figure*}

\section{Implementation Details} \label{network}

\vspace{2mm} \noindent \textbf{Dataset Details.} We conduct our experiments on two categories of data: monocular images of human portraits and cat faces. We follow the method in EG3D~\cite{chan2022efficient} to extract the camera parameters of these images with off-the-shelf pose detectors ~\cite{deng2019accurate, cat_hipsterizer}. For human portraits, we use FFHQ~\cite{abdal2019image2stylegan} which contains about 70,000 images to train our model. To evaluate our model's performance of input-view reconstruction, we randomly sample 1,500 images from CelebA-HQ~\cite{karras2018progressive} test for quantitative evaluation. Additionally, we use a multi-view dataset MEAD~\cite{wang2020mead} to evaluate our model's performance across novel views. Specifically, we use five views (left$60^{\circ}$, left$30^{\circ}$, front, right$30^{\circ}$, right$60^{\circ}$) images of 43 persons per frame. We randomly sample 5 frames for each person. For cat faces, we use AFHQv2 Cats~\cite{choi2020stargan} following EG3D. We split about 5,000 images into train, evaluation, and test sets by 8:1:1 ratio.

\vspace{2mm} \noindent \textbf{Architecture of Geometry-aware Encoder.} Our encoder uses Swin-transformer as the backbone, and we further design attention modules at different scale feature layers for different level latent codes. The encoder architecture is shown in Fig.~\ref{Eswin}. We split the intermediate output of the Swin-transformer into four levels ``query, coarse, mid, fine"   similar to the pyramid architecture of CNN models. We use ``query" to get $w_0$ and query $Q_{coarse}, Q_{mid}, Q_{fine}$, and leverage ``coarse", ``mid", ``fine" to obtain keys and values $(K,V)_{coarse}, (K,V)_{mid}, (K,V)_{fine}$. Then the different level queries with corresponding keys and values are sent into cross-attention modules to yield different $w_i$. Finally, the final latent code $w^+$ is obtained by:

\begin{equation}
\begin{aligned}
    w^+ = w_{avg} + w_0 + (0, w_{1\sim3}, w_{3\sim6}, w_{7\sim13}).
\end{aligned}
\end{equation}

\vspace{2mm} \noindent \textbf{Occlusion-aware Mix Tri-plane.} As mentioned in Sec.4.3, We can get the visible points set $\mathcal{V}_{(x,y,z)}$. Then we perform orthogonal projection that projects these points to the three axis-aligned feature planes ($F_{xy}$, $F_{xz}$, $F_{yz}$) of tri-plane to get three masks separately, denotes as $tri\text{-}Mask$. The grid point in $tri\text{-}Mask$ is equal to 1 if the corresponding 3D point is in $\mathcal{V}_{(x,y,z)}$, otherwise it is equal to 0. Finally, the mix tri-plane can be obtained by:
\begin{equation}
\begin{aligned}
tri\text{-}plane_{mix} &= tri\text{-}plane_{F^*} \odot tri\text{-}Mask \\
&+ tri\text{-}plane_{w^+} \odot (tri\text{-}I-tri\text{-}Mask), \\    
\end{aligned}
\end{equation}
where $tri\text{-}I$ is the concatenated result of three all-one matrices, which has the same dimension as $tri\text{-}Mask$.

\vspace{2mm} \noindent \textbf{Training Strategy} 
We use a two-stage training strategy. In the first stage, we only train our encoder model based on the loss in Sec.~4.1. After the loss is converged, we freeze the encoder parameters, and train the Adaptive Feature Alignment (AFA) module with occlusion-aware mix tri-plane. When training the geometry-aware encoder, we follow e4e~\cite{tov2021designing} to train different level modules and output corresponding $w_i$ progressively. Only the latent code before and at the current stage will be added to $w^+$. Different from e4e, we only use 3 progressives stage (\ie coarse, mid, fine), instead of 14 $w_i$ stages (\ie $0, 1, 2, ..., 13$) in e4e. We train the canonical latent discriminator $\mathcal{D}_{\mathcal{W}_c}$ at the beginning of each encoder training iteration, and fix its parameter when training the encoder.

\vspace{2mm} \noindent \textbf{Losses. }In the first training stage, we use the loss function mentioned in Sec.~4.1. For the input view reconstruction, the loss $\mathcal{L}_{rec}$ is computed by:
\begin{equation}
    \begin{footnotesize}
    \begin{aligned}
        \mathcal{L}_{rec} &= \lambda_1 \left \| I_{w^+} -  I \right \|_2  + \lambda_2 \mathcal{L}_{LPIPS}(I_{w^+}, I) + \lambda_3\mathcal{L}_{ID}(I_{w^+},I),
    \end{aligned}
    \end{footnotesize}
\end{equation}
where $\lambda_1 = 1.0, \lambda_2 = 0.8, \lambda_3 = 0.25$ are constant weights.
The total loss for training our encoder and the canonical latent discriminator can be written as:
\begin{equation}
    \mathcal{L}^E_{total} = \mathcal{L}^\mathcal{D}_{\text{adv}} + \lambda_4\mathcal{L}_{\text{adv}}^E   + \lambda_5\mathcal{L}_{BG} + \mathcal{L}_{rec},
\end{equation}
where $R_1$ regularization weight $\gamma=10$, $\lambda_4 = 0.05, \lambda_5 = 5.0$ are constant weights. Moreover, we apply regularization to $\Delta w^+ = (w_1-w_0, w_2-w_0, ..., w_{13}-w_0)$ by $\mathcal{L}_2$ norm, \ie $\mathcal{L}_{w^+} = \lambda_6\left \| \Delta w^+ \right \|_2 $, where $\lambda_6 = 0.001$. 

Once the loss in the first stage is converged, we freeze the encoder parameters and drop the canonical latent discriminator and losses $\mathcal{L}^\mathcal{D}_{\text{adv}}, \mathcal{L}_{\text{adv}}^E, \mathcal{L}_{w^+}$, but add a $\mathcal{L}_2$ regularization loss on $\Delta F = F^*-F$ with $\lambda_7=0.0001$ for training the AFA module. We apply the loss settings to both human data and cat data.

For re-implemented pSp~\cite{richardson2021encoding} and e4e~\cite{tov2021designing}, we train it with the same loss weight in our settings, and train them until the losses are converged. 

\vspace{2mm} \noindent \textbf{Experiment settings. }For human data, we train the encoder with a batch size of 3 on one 3090 GPU for about 1,000,000 iterations. We use a discriminator learning rate of 0.00002 with Adam optimizer and an encoder learning rate of 0.0001 with a ranger optimizer, which is a combination of Rectified Adam ~\cite{liu2019variance} with the Lookahead technique~\cite{zhang2019lookahead}. Then we train our AFA module with a learning rate of 0.000025 with Adam optimizer, which uses 800,000 iterations of batch size 2. We only train the encoder 200,000 iterations and 150,000 iterations for cat data. We test our method and other methods with 3090 GPUs, and test inference time with the same settings. We use images from CelebA-HQ or AFHQv2 Cats to perform inversion.

\vspace{2mm} \noindent \textbf{IDE-3D. }IDE-3D~\cite{sun2022ide} uses semantic segmentation to re-train EG3D. It learns an encoder to get a latent code and further optimize the generator parameters for inversion and editing images. We compare our encoder with its encoder both qualitatively and quantitatively.

\begin{table*}[h] 
	\renewcommand\arraystretch{1.05}
	\centering  
	\caption{\textbf{Ablation studies on Geometry-aware Encoder and Occlusion-aware Mix Tri-plane.} We evaluate the proposed canonical space $\mathcal{W}_c$, the background regularization, and the occlusion-aware mix tri-plane on MEAD dataset. The best performance on $\mathcal{W}$ space inversion and $\mathcal{F}$ space inversion are in bold.} 
	\label{novel view metrics}  
	\resizebox{1.0\linewidth}{!}{
    \begin{tabular}{c|ccccccc|ccccccc}
    \toprule  
    \multirow{3}{*}{ Method } & \multicolumn{14}{c}{ Novel view quality }   \\
    \cline{2-15} & \multicolumn{7}{|c|}{ yaw$=\pm60^{\circ}$  } & \multicolumn{7}{|c}{ yaw $=0^{\circ}$ }  \\
    \cline{2-15} & PSNR $\uparrow$ & SSIM $\uparrow$ & MSE $\downarrow$& LPIPS $\downarrow$ & $\mathrm{FID} \downarrow $ & $\mathrm{ID} \uparrow$ & Geo. Err. $\downarrow$ & PSNR $\uparrow$ & SSIM $\uparrow$ & MSE $\downarrow$ & LPIPS $\downarrow$ & $\mathrm{FID} \downarrow$ & $\mathrm{ID} \uparrow$ & Geo. Err. $\downarrow$ \\ \hline
    w/o $\mathcal{D}_{\mathcal{W}_c}$, w/o $\mathcal{L}_{BG}$    &19.82 &0.5730 &0.0565 &0.3069 &62.1 &0.6206 &0.1422 &19.83 &0.5382  &0.0540&0.2966 &68.9 &0.6842 &0.1775 \\
    w/o $\mathcal{D}_{\mathcal{W}_c}$, w/ $\mathcal{L}_{BG}$    &19.41 &0.5576 &0.0585 &0.2883 &59.0 &0.6176 &0.1016 &20.76 &0.5829 &0.0417 &0.2572 &60.6 &\textbf{0.6864} &0.1082\\
    w/ $\mathcal{D}_{\mathcal{W}_c}$, w/o $\mathcal{L}_{BG}$  &20.20 &0.5999 &0.0495 &0.2599 &55.5 &0.6459 &0.1414 &20.67 &0.5757 &0.0428 &0.2527 &55.1 &0.6844 &0.1491\\
    w/ $\mathcal{D}_{\mathcal{W}_c}$, w/ $\mathcal{L}_{BG}$  &\textbf{20.66} &\textbf{0.6211} &\textbf{0.0473} &\textbf{0.2203} &\textbf{52.2} &\textbf{0.6552} &\textbf{0.0943} &\textbf{21.12} &\textbf{0.5944} &\textbf{0.0383} &\textbf{0.2372} &\textbf{54.1} &0.6848 &\textbf{0.0955}\\
    \hline
     w/o occlusion-aware &20.69 &0.6176 &0.0489 &0.2327 &52.5 &0.6832 &0.0983 &20.26 &0.6374 &\textbf{0.03202} &0.2230 &50.3 &\textbf{0.744}  &0.0954   \\
     Ours & \textbf{20.87}	& \textbf{0.6299} & \textbf{0.0424} & \textbf{0.2192} & \textbf{50.9} & \textbf{0.700} & \textbf{0.0950} & \textbf{22.13}	& \textbf{0.6502}  & 0.0329 & \textbf{0.2204} & \textbf{51.7} & 0.743 & \textbf{0.0944} \\
    \bottomrule
\end{tabular}}
\end{table*}

\begin{table}[h] 
    \renewcommand\arraystretch{1.05}
    \centering  
    \caption{\textbf{Ablation study on AFA module.} We replace the proposed adaptive alignment module in AFA with CNN network and evaluate on CelebA-HQ dataset.} 
    \label{Input view Inversion metrics}  
    \setlength{\tabcolsep}{6 pt}
    \resizebox{1.0\linewidth}{!}{
    \begin{tabular}{l|ccccccc}
        \toprule 
        \multirow{1}{*}{ Method }  & PSNR $\uparrow$ & SSIM $\uparrow$ & MSE $\downarrow$  &  LPIPS $\downarrow$ & FID $\downarrow$ &  ID $\uparrow$ & Geo. Err. $\downarrow$  \\
            \hline 
            \multirow{1}{*}{w/ conv} &19.67 &0.5962 &0.0455 &0.1766 &25.1 &0.8033  &0.1094  \\
            \multirow{1}{*}{w/ AFA}  &\textbf{21.84} &\textbf{0.7079} &\textbf{0.301} &\textbf{0.1242} &\textbf{18.1} &\textbf{0.8797}  &\textbf{0.0984}  \\

            \bottomrule
    \end{tabular}}
\end{table}

\vspace{2mm} \noindent \textbf{PTI. }PTI~\cite{roich2022pivotal} is the most used optimization method in 2D GAN inversion. We follow its official settings, and only replace the 2D GAN by the EG3D generator. It learns the pivot latent code for 450 iterations, and finetunes generator parameters for 350 iterations.

\vspace{2mm} \noindent \textbf{Pose Opt. }Pose Opt.~\cite{ko20223d} jointly optimizes camera pose, latent codes, and generator parameters for 3D GAN inversion. We follow its official settings and learn the pivot latent code and camera pose for 400 iterations at the first stage, and finetune generator parameters for 400 iterations at the second stage. We also use its pre-trained encoder for pivot latent code and camera pose initialization.

\vspace{2mm} \noindent \textbf{Editing. }We perform InterfaceGAN~\cite{shen2020interpreting} to get a semantic latent direction for editing. First, we sample 500000 front faces whose corresponding latent codes are conditioned by canonical poses, and sort them by attribution classifiers. We choose the top 10000 and bottom 10000 samples according to their score of classification, then we use SVM to get the direction. Finally, We use the method in Sec.~4.4 to get 3D-consistent editing results.

\section{More Experimental Results} \label{results}
\vspace{2mm} \noindent \textbf{Ablation of Geometry-aware Encoder Designs.} We test different ablation settings of our encoder on MEAD, for novel view geometry and texture evaluation. As shown in Table~\ref{novel view metrics}, the design of the canonical discriminator ($\mathcal{D}_{\mathcal{W}_c}$) and background depth regularization ($\mathcal{L}_{BG}$) is necessary for a good geometry inversion.

\vspace{2mm} \noindent \textbf{Ablation of Adaptive Feature Alignment. }We evaluate Adaptive Feature Alignment (AFA) module ablation on source view reconstruction performance on CelebA-HQ, as shown in Table ~\ref{Input view Inversion metrics}. Modified feature maps generated by only convolution modules are hard to align to the facial region in canonical feature space $\mathcal{F}_c$, whose reconstruction quality is inferior to our method.

\vspace{2mm} \noindent \textbf{Ablation of Occlusion-aware Mix Tri-plane. }We evaluate occlusion-aware mix tri-plane design on MEAD for novel view evaluation. As the distortion exists in the occlusion part, our full models will perform better on novel view image synthesis, as shown in Table ~\ref{novel view metrics}.

\begin{figure*}[t]
	\begin{center}
		\vspace{-0.2in}
		\includegraphics[width=1\linewidth]{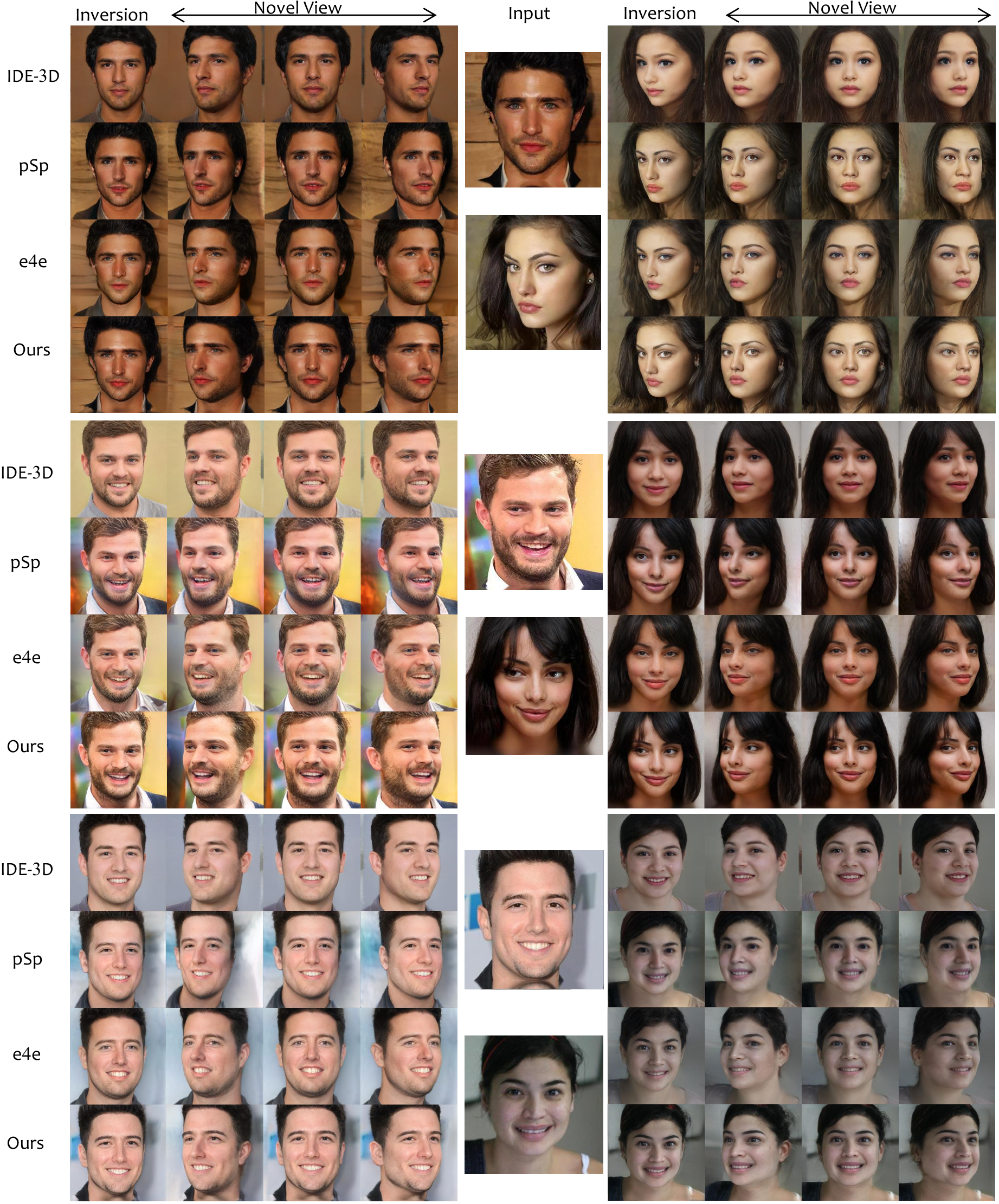}
	\end{center}
	\vspace{-0.6cm}
	\caption{\textbf{Comparison with encode-based methods.}}
	\label{compare_encoder}
	\vspace{-0.3cm}
\end{figure*}

\vspace{2mm} \noindent \textbf{Comparison with Encode-based Methods.} We show more qualitative comparisons with encode-based methods in Fig.~\ref{compare_encoder}. Our method significantly surpasses others.

\begin{figure*}[t]
	\begin{center}
		\vspace{-0.2in}
		\includegraphics[width=1\linewidth]{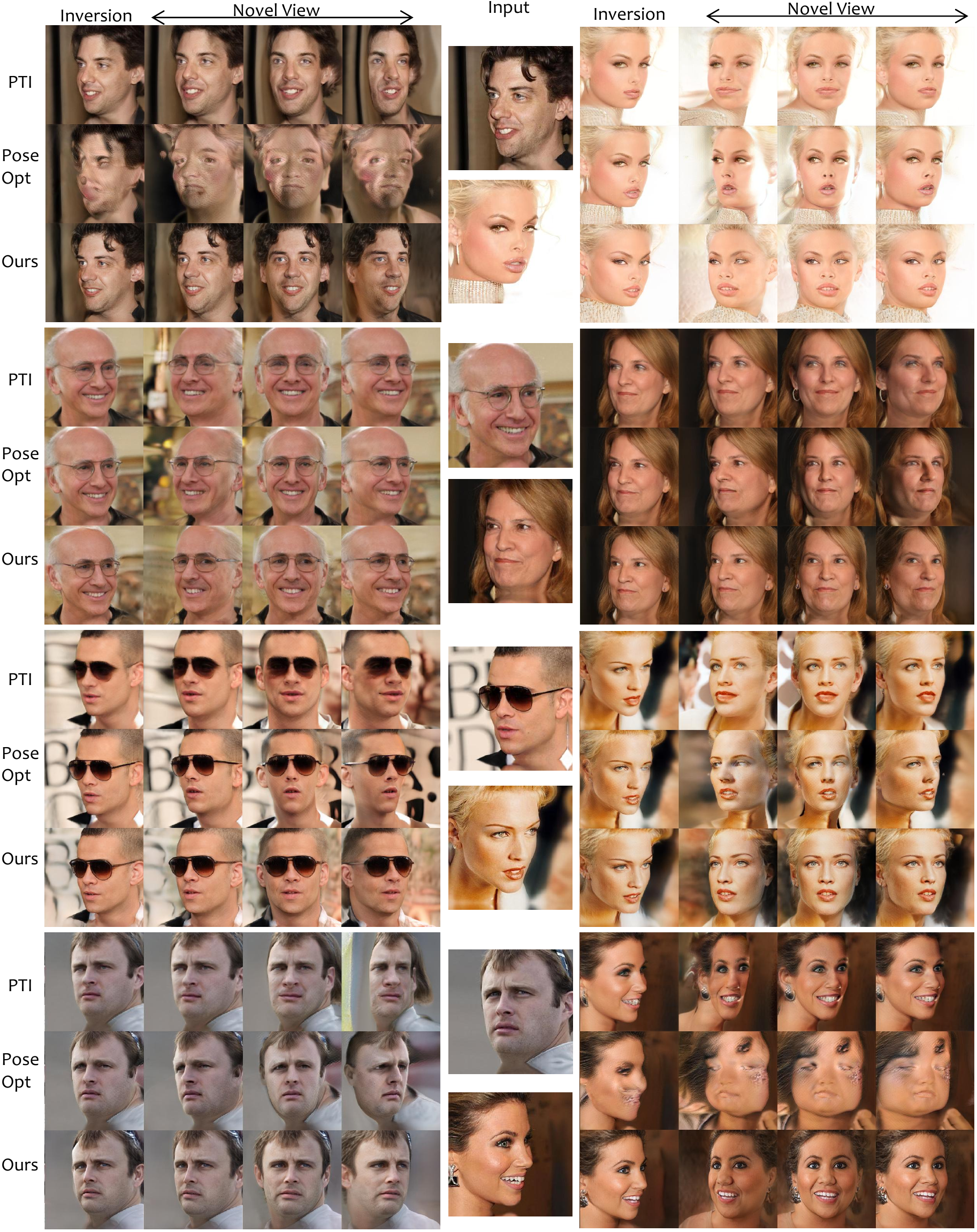}
	\end{center}
	\vspace{-0.6cm}
	\caption{\textbf{Comparison with optimization-based methods.}}
	\label{compare_optimization}
	\vspace{-0.3cm}
\end{figure*}

\vspace{2mm} \noindent \textbf{Comparison with Optimization-based Methods. }We present more qualitative results of comparison with optimization-based methods in Fig.~\ref{compare_optimization}. It is worth noting that when the input image is a side face, optimization-based methods tend to overfit the input image and are hard to synthesize novel-view images. In some cases, Pose Opt. fails to converge to an accurate pose using optimization and generates degradation results.

\begin{figure*}[t]
	\begin{center}
		\vspace{-0.2in}
		\includegraphics[width=1\linewidth]{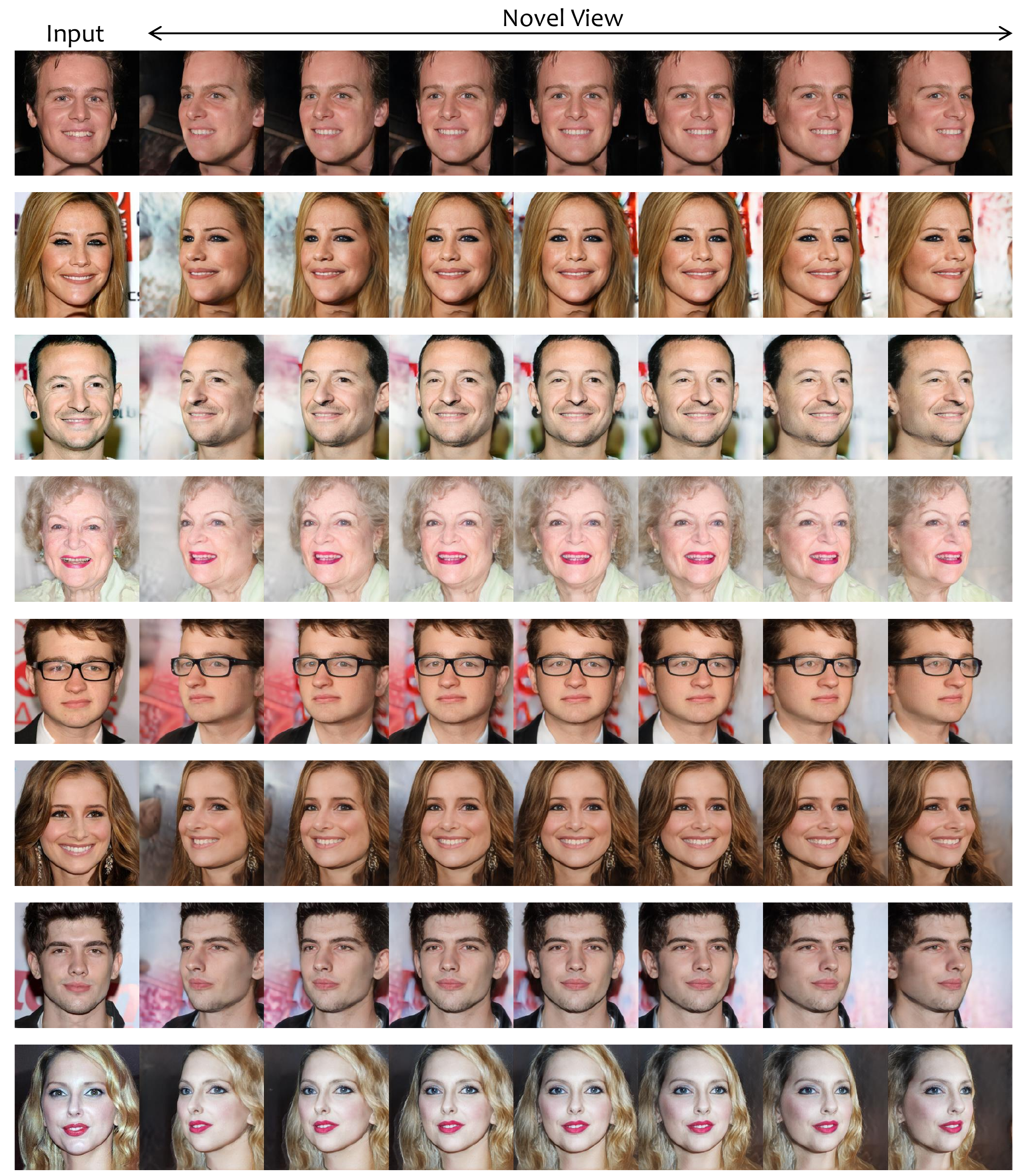}
	\end{center}
	\vspace{-0.6cm}
	\caption{\textbf{Multi-view human faces inversion results of our method (part 1/2).}}	
	\label{novel_view1}
	\vspace{-0.3cm}
\end{figure*}

\begin{figure*}[t]
	\begin{center}
		\vspace{-0.2in}
		\includegraphics[width=1\linewidth]{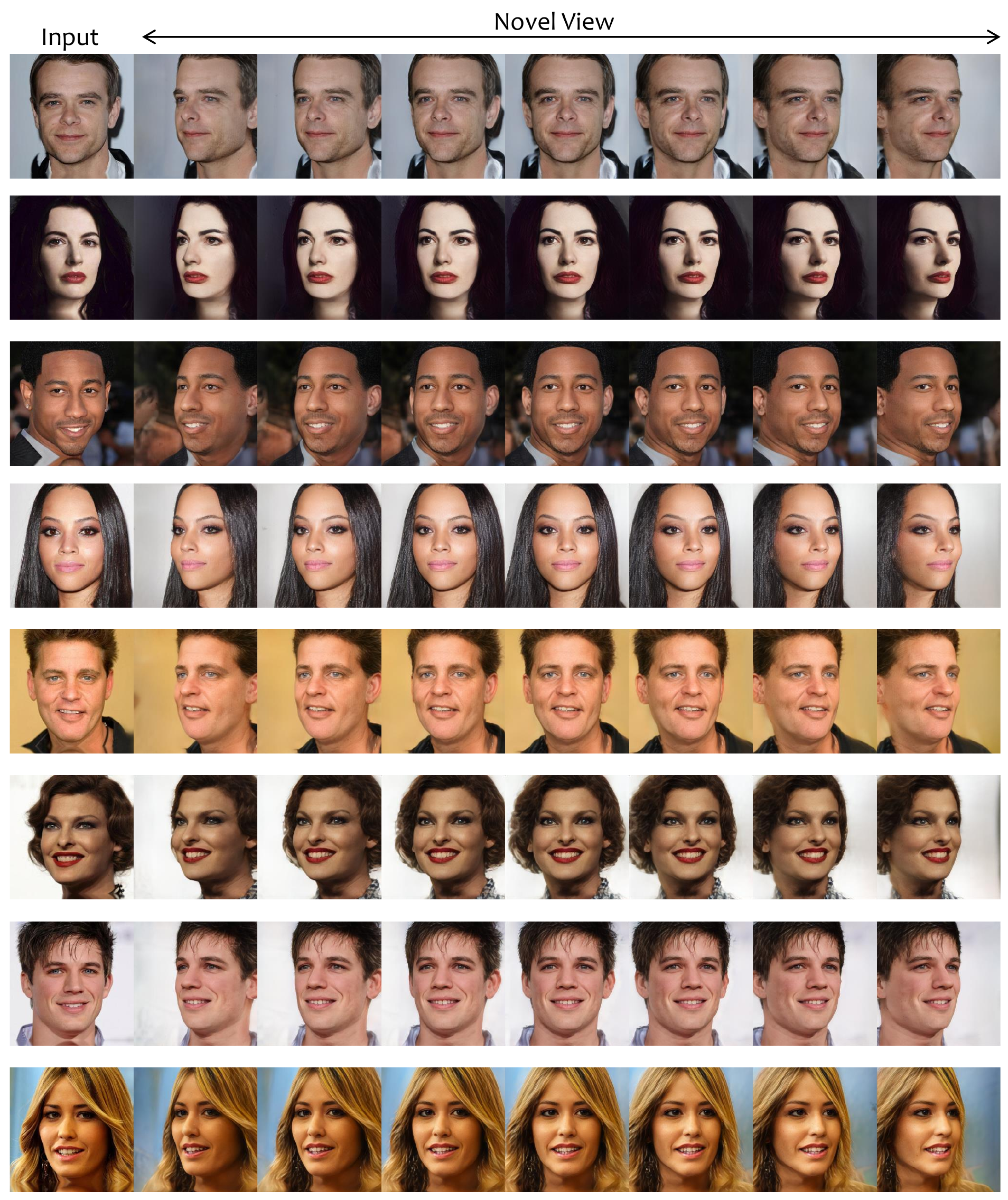}
	\end{center}
	\vspace{-0.6cm}
	\caption{\textbf{Multi-view human faces inversion results of our method (part 2/2).}}
	\label{novel_view2}
	\vspace{-0.3cm}
\end{figure*}

\begin{figure*}[t]
	\begin{center}
		\vspace{-0.2in}
		\includegraphics[width=1\linewidth]{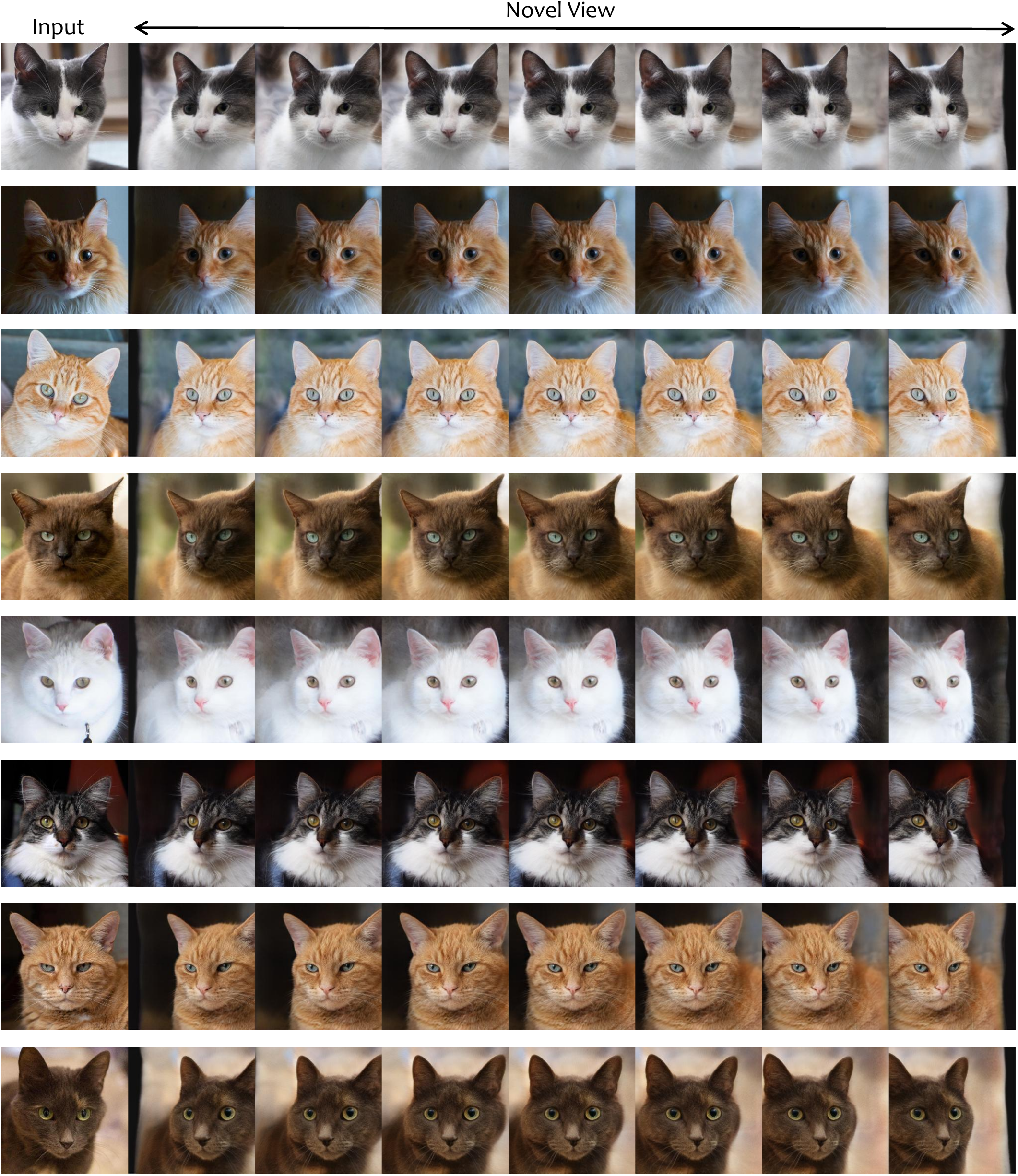}
	\end{center}
	\vspace{-0.6cm}
	\caption{\textbf{Multi-view cat faces inversion results of our method.}}
	\label{cat_novel_view}
	\vspace{-0.3cm}
\end{figure*}

\vspace{2mm} \noindent \textbf{More Novel View Results with Our Method. }  More novel view results with our method can be found in Fig.~\ref{novel_view1},~\ref{novel_view2},~\ref{cat_novel_view}. Our method can achieve high-quality 3D-consistent multi-view image synthesis.

\begin{figure*}[t]
	\begin{center}
		\vspace{-0.2in}
		\includegraphics[width=1\linewidth]{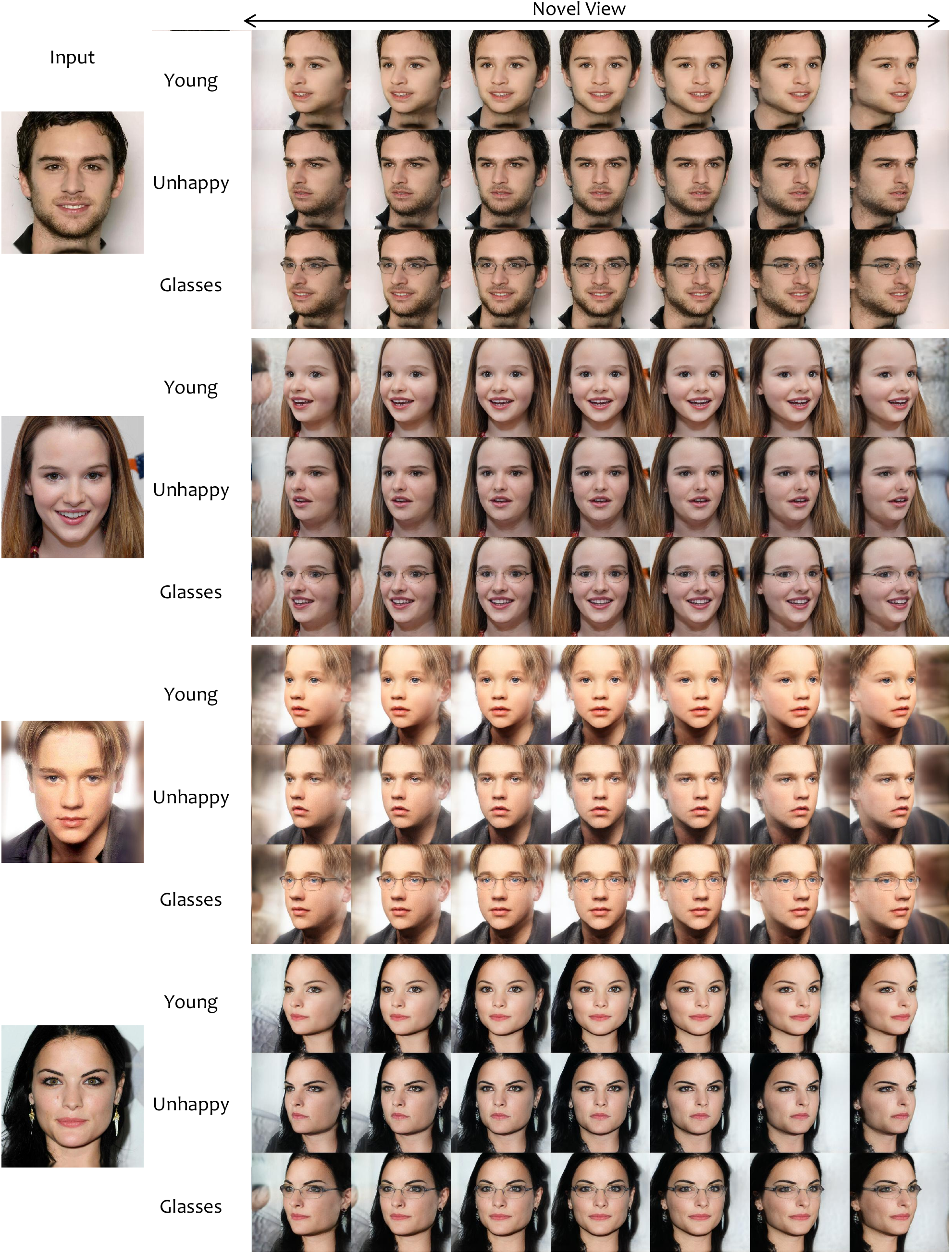}
	\end{center}
	\vspace{-0.6cm}
	\caption{\textbf{Editing results on human faces (part 1/2).}}
	\label{edit_human1}
	\vspace{-0.3cm}
\end{figure*}

\begin{figure*}[t]
	\begin{center}
		\vspace{-0.2in}
		\includegraphics[width=1\linewidth]{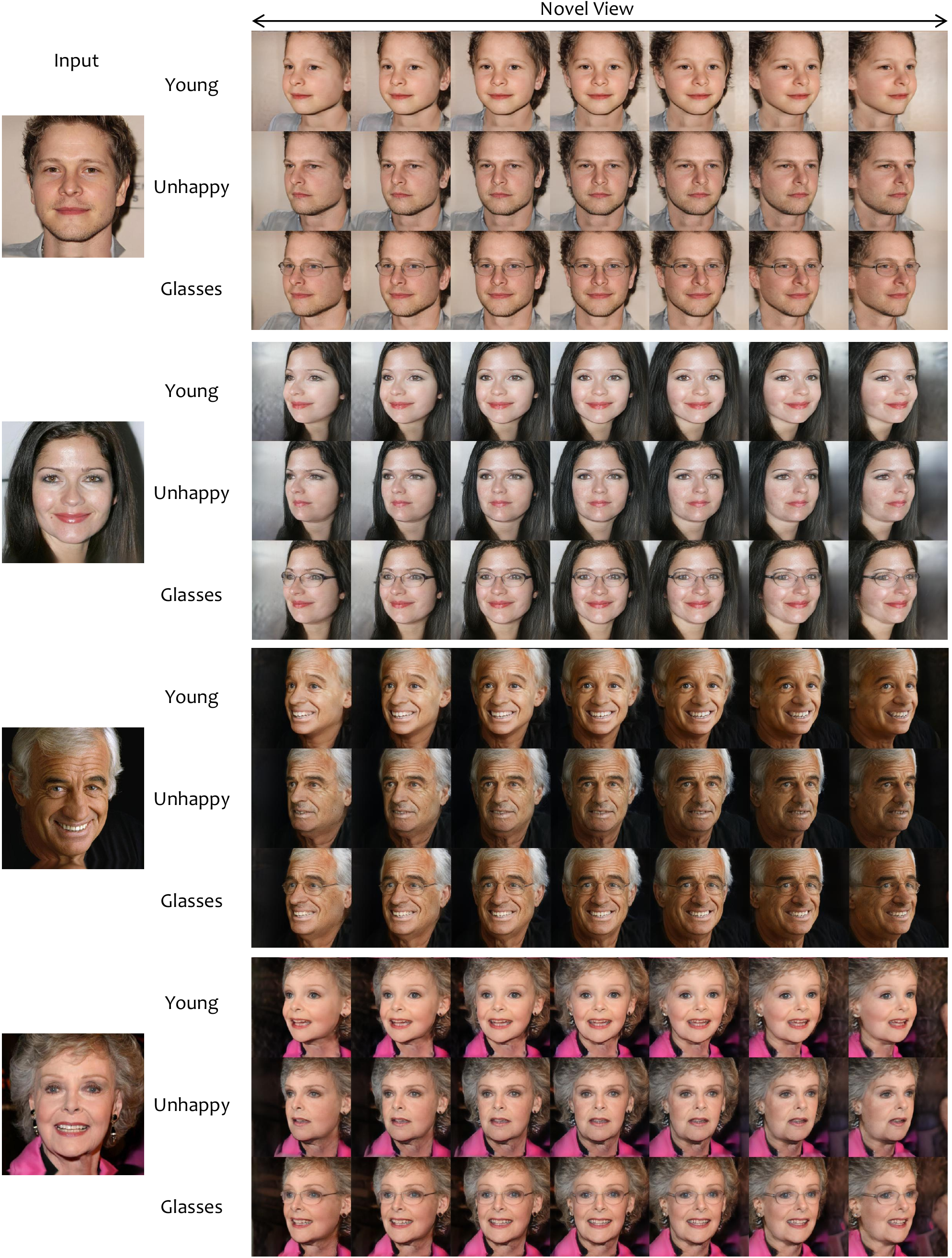}
	\end{center}
	\vspace{-0.6cm}
	\caption{\textbf{Editing results on human faces (part 2/2).}}
	\label{edit_human2}
	\vspace{-0.3cm}
\end{figure*}

\begin{figure*}[t]
	\begin{center}
		\vspace{-0.2in}
		\includegraphics[width=1\linewidth]{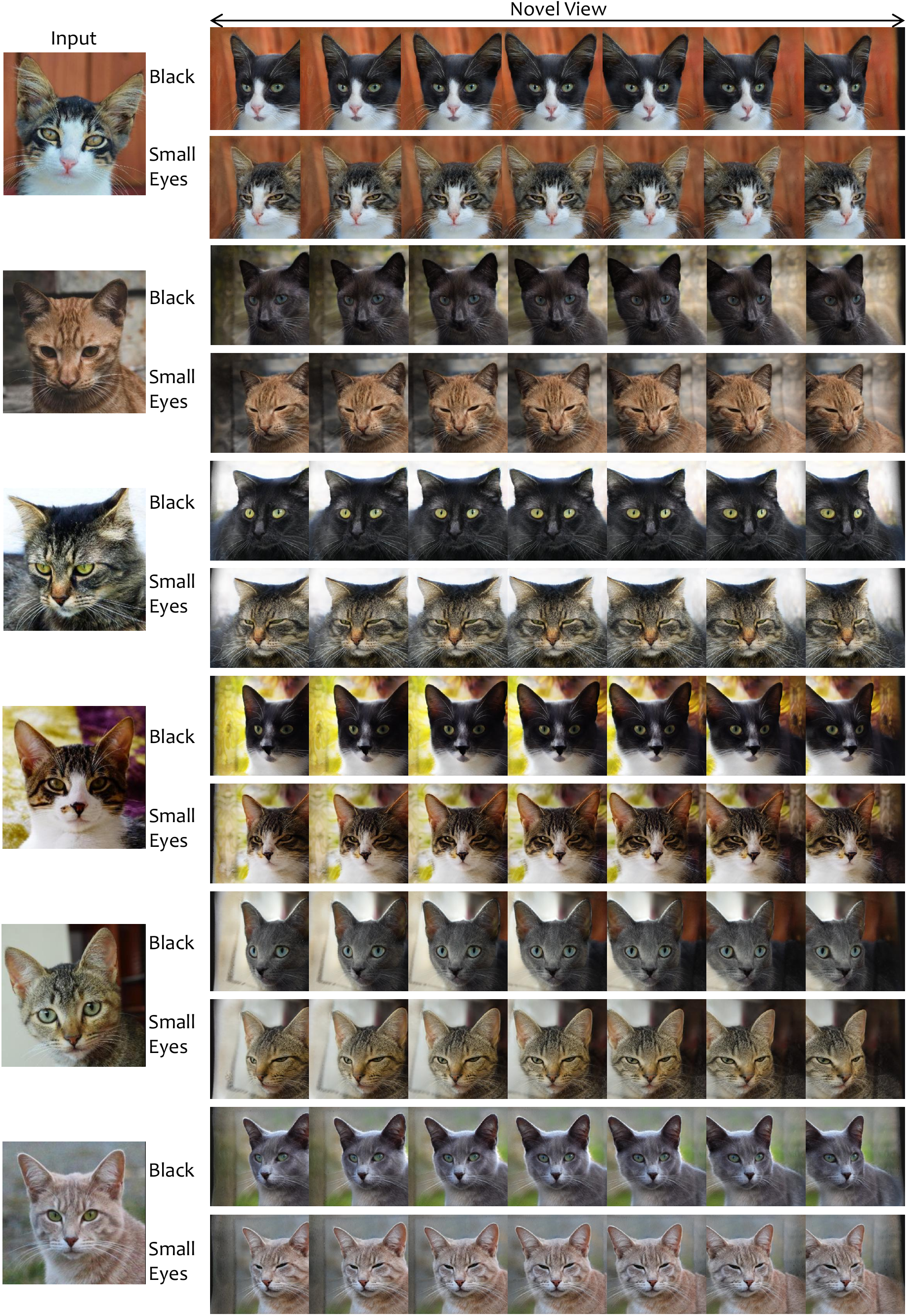}
	\end{center}
	\vspace{-0.6cm}
	\caption{\textbf{Editing results on cat faces.}}
	\label{edit_cat}
	\vspace{-0.3cm}
\end{figure*}

\vspace{2mm} \noindent \textbf{More Results of Editing with Our Method. }We present more 3D-consistent results of human faces and cat faces in Fig.~\ref{edit_human1},~\ref{edit_human2},~\ref{edit_cat}. Our method shows powerful editing ability which can be used in real-world applications.


\end{document}